\documentclass[runningheads]{llncs}
\usepackage{graphicx}

\usepackage{tikz}
\usepackage{comment}
\usepackage{amsmath,amssymb} 
\usepackage{color}

\usepackage{comment}

\usepackage[accsupp]{axessibility}  

\usepackage[width=122mm,left=12mm,paperwidth=146mm,height=193mm,top=12mm,paperheight=217mm]{geometry}

\usepackage{multirow}

\usepackage{booktabs} 
\usepackage{colortbl} 
\usepackage{graphicx} 
\usepackage{subfigure} 
\usepackage{enumitem}  
\usepackage{xspace}  
\usepackage[bottom]{footmisc}   

\makeatletter
\DeclareRobustCommand\onedot{\futurelet\@let@token\@onedot}
\def\@onedot{\ifx\@let@token.\else.\null\fi\xspace}

\def\eg{\emph{e.g}\onedot} 
\def\ie{\emph{i.e}\onedot} 
 
\def\etc{\emph{etc}\onedot} \def\vs{\emph{vs}\onedot}
\def\wrt{w.r.t\onedot}

\makeatother

\let\vec\mathbf 
\newcommand{\R}[0]{\mathbb{R}}
\newcommand{\N}[0]{\mathbb{N}}
\newcommand{\eqcomma}[0]{\;\;,}
\newcommand{\eqdot}[0]{\;\;.}

\newcommand{\train}{{\small\texttt{train}\xspace}}
\newcommand{\val}{{\small\texttt{val}\xspace}}
\newcommand{\test}{{\small\texttt{test}\xspace}}

\newcommand{\visualprompt}[0]{Visual-Prompt Tuning}
\newcommand{\vprompt}[0]{\textsc{VPT}}
\newcommand{\deepprompt}[0]{\vprompt{}-\textsc{deep}}
\newcommand{\shallowprompt}[0]{\vprompt{}-\textsc{shallow}}

\newcommand{\partialft}[0]{\textsc{Partial}}
\newcommand{\linear}[0]{\textsc{Linear}}
\newcommand{\fullft}[0]{\textsc{Full}}
\newcommand{\sidetune}[0]{\textsc{Sidetune}}
\newcommand{\mlp}[0]{\textsc{Mlp}}
\newcommand{\bias}[0]{\textsc{Bias}}
\newcommand{\adapter}[0]{\textsc{Adapter}}
\newcommand{\promptbias}[0]{\textsc{VPT+Bias}}

\newcommand{\vit}[0]{ViT}
\newcommand{\rn}[0]{ResNet}
\newcommand{\swin}[0]{Swin}
\newcommand{\rnx}[0]{ConvNeXt}

\newcommand{\suplong}[0]{Supervised}

\newcommand{\moco}[0]{MoCo v3}

\newcommand{\mae}[0]{MAE}

\newcommand{\imagenet}[0]{ImageNet}

\newcommand{\cub}[0]{CUB-200-2011}
\newcommand{\nabirds}[0]{NABirds}
\newcommand{\flowers}[0]{Oxford Flowers}
\newcommand{\cars}[0]{Stanford Cars}
\newcommand{\dogs}[0]{Stanford Dogs}
\newcommand{\vtab}[0]{VTAB-1k}

\definecolor{tabvline}{HTML}{a8a495}
\definecolor{prompt_blue}{HTML}{1f78b4}
\definecolor{prompt_red}{HTML}{d45c43}

\definecolor{green_im}{rgb}{0.0, 0.5, 0.0}
\newcommand{\ttbf}[1]{\textbf{\texttt{#1}}}
\newcommand{\band}{\rowcolor{gray!15}}

\newcommand{\Drop}[1]{\textcolor{prompt_red}{\xspace\scriptsize{\bf $\downarrow$#1}}}
\newcommand{\Rise}[1]{\textcolor{green_im}{\xspace\scriptsize{\bf $\uparrow$#1}}}

\newcommand{\para}[1]{\subsubsection{#1}}

\definecolor{citecolor}{RGB}{0, 113, 188}

\usepackage[pagebackref,breaklinks,colorlinks,citecolor=citecolor]{hyperref}

\usepackage[capitalize]{cleveref}
\crefname{section}{Sec.}{Secs.}
\Crefname{section}{Section}{Sections}
\Crefname{table}{Table}{Tables}
\crefname{table}{Tab.}{Tabs.}

\makeatletter
\newcommand{\printfnsymbol}[1]{%
  \textsuperscript{\@fnsymbol{#1}}%
}
\makeatother

\begin{document}
\pagestyle{headings}
\mainmatter

\title{Visual Prompt Tuning}

\titlerunning{Visual Prompt Tuning}
%

\author{Menglin Jia\thanks{Equal contribution.}$^{1,2}$
\and
Luming Tang\printfnsymbol{1}$^{1}$
\\
Bor-Chun Chen$^{2}$
\and
Claire Cardie$^{1}$
\and
Serge Belongie$^{3}$
\\
Bharath Hariharan$^{1}$
\and
Ser-Nam Lim$^{2}$
}

\authorrunning{M. Jia et al.}
%
\institute{$^{1}$Cornell University \qquad $^{2}$Meta AI \qquad $^{3}$University of Copenhagen}
\maketitle

\begin{abstract}
The current \textit{modus operandi} in adapting pre-trained models involves updating all the backbone parameters, \ie, full fine-tuning.
This paper introduces Visual Prompt Tuning (VPT) as an efficient and effective alternative to full fine-tuning for large-scale Transformer models in vision.
Taking inspiration from recent advances in efficiently tuning large language models, 
VPT introduces only a small amount (less than 1\% of model parameters) of trainable parameters in the input space while keeping the model backbone frozen.
Via extensive experiments on a wide variety of downstream recognition tasks, we show that VPT achieves significant performance gains compared to other parameter efficient tuning protocols. Most importantly, VPT even outperforms full fine-tuning in many cases across model capacities and training data scales, while reducing per-task storage cost. 
Code is available at \href{https://github.com/kmnp/vpt}{\texttt{github.com/kmnp/vpt}}.
\end{abstract}

\section{Introduction}\label{sec: intro_luming}

\begin{figure}[t]
\centering
\includegraphics[width=\textwidth]{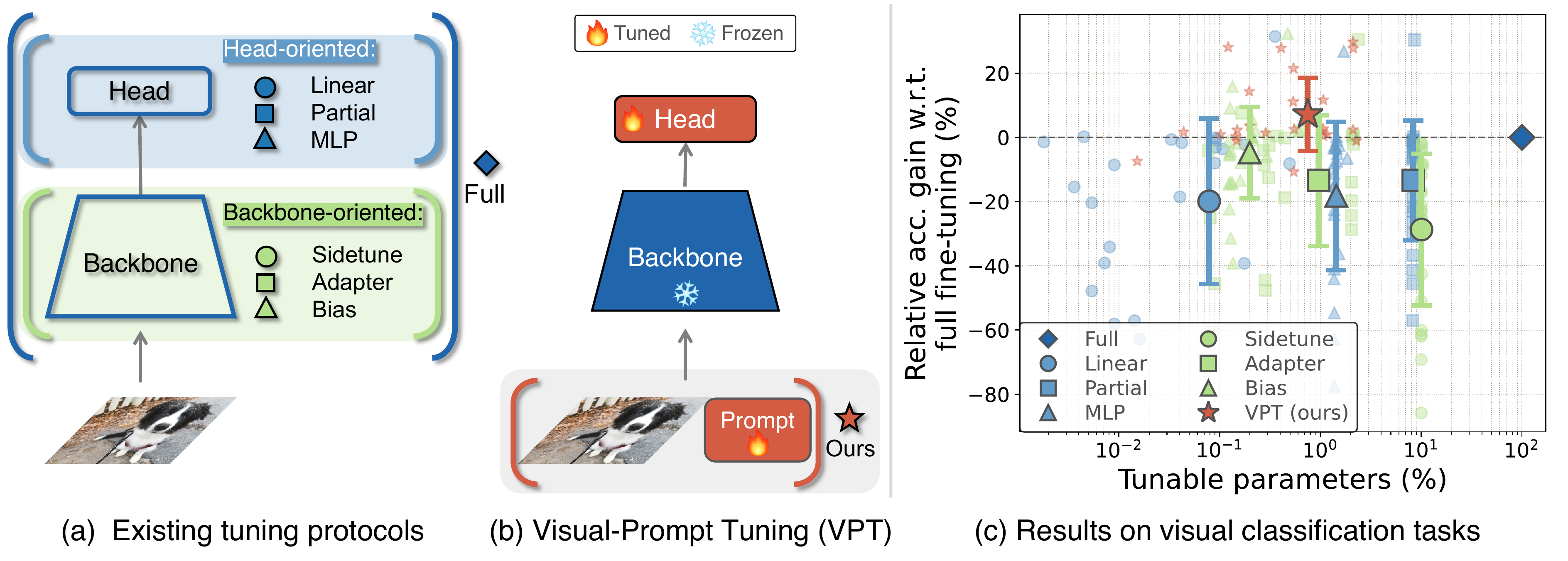}
\vspace{-0.7cm}
\caption{
Visual-Prompt Tuning (VPT) \vs~other transfer learning methods. 
(a) Current transfer learning protocols are grouped based on the tuning scope: Full fine-tuning, Head-oriented, and Backbone-oriented approaches.
(b) VPT instead adds extra parameters in the input space. 
(c) Performance of different methods on a wide range of downstream classification tasks adapting a pre-trained \vit{}-B backbone, with mean and standard deviation annotated.
VPT outperforms Full fine-tuning 20 out of 24 cases while using less than 1$\%$ of all model parameters
}
\vspace{-0.5cm}
\label{fig:teaser}
\end{figure}

For a variety of recognition applications, the most accurate results are now obtained by adapting large \emph{foundation models} pre-trained on massive curated or raw data, a finding that mirrors developments in natural language processing (NLP)~\cite{bommasani2021opportunities}.\footnote{As pointed out in~\cite{bommasani2021opportunities}, all state-of-the-art models in contemporary NLP are now powered by a few Transformer-based models (\eg, BERT~\cite{devlin-etal-2019-bert}, T5~\cite{2020t5}, BART~\cite{lewis2020bart}, GPT-3~\cite{brown2020gpt3}) This also applies to vision-language field recently, \ie, CLIP~\cite{radford2021learning}.}
At first glance,this is a success story: one can make rapid progress on multiple recognition problems simply by leveraging the latest and greatest foundation model.
In practice, however, \emph{adapting} these large models to downstream tasks presents its own challenges.
The most obvious (and often the most effective) adaptation strategy is \emph{full fine-tuning} of the pre-trained model on the task at hand, end-to-end.
However, this strategy requires one to store and deploy a separate copy of the backbone parameters for every single task.
This is an expensive and often infeasible proposition, especially for modern \emph{Transformer}-based architectures, which are significantly larger than their convolutional neural networks (ConvNet) counterparts, \eg, ViT-Huge~\cite{dosovitskiy2020vit} (632M parameters) \vs~ResNet-50~\cite{he2016rn} (25M parameters).
We therefore ask, \textbf{what is the best way to adapt large pre-trained Transformers to downstream tasks in terms of effectiveness and efficiency}?

One straightforward approach is to turn to other strategies that we have perfected for adapting ConvNets to new tasks, as in~\cref{fig:teaser}(a).
A popular approach is to fine-tune only a subset of the parameters, such as the classifier head~\cite{wslimageseccv2018,jia2021exploring,chen2021mocov3} or the bias terms~\cite{cai2020tinytl}. 
Prior research has also looked at adding additional residual blocks  (or \emph{adapters}) to the backbone~\cite{rebuffi2018efficient,zhang2020side}.
One could implement similar strategies for Transformers.
However, in general these strategies \emph{under-perform} full fine-tuning in accuracy.

We explore a different route in this paper. Instead of altering or fine-tuning the pre-trained Transformer itself, we modify the \emph{input} to the Transformer.
Drawing inspiration from the recent advances on Prompting in NLP~\cite{liu2021pre,li-liang-2021-prefix,lester-etal-2021-power,liu2021p}, we propose a new simple and efficient method to adapt transformer models for downstream vision tasks (\cref{fig:teaser}(b)), namely \textbf{Visual-Prompt Tuning} (VPT). Our method only introduces a small amount of task-specific learnable parameters into the input space while freezing the entire pre-trained Transformer backbone during downstream training. 
In practice, these additional parameters are simply prepended into the input sequence of each Transformer layer and learned together with a linear head during fine-tuning.

On 24 downstream recognition tasks spanning different domains using a pre-trained \vit{} backbone, VPT beats all other transfer learning baselines, even surpassing full fine-tuning in 20 cases, while maintaining the advantage of storing remarkably fewer parameters (less than 1\% of backbone parameters) for each individual task (\cref{fig:teaser}(c)). 
This result demonstrates the distinctive strength of \emph{visual} prompting: whereas in NLP, prompt tuning is only able to \emph{match} full fine-tuning performance under certain circumstances~\cite{lester-etal-2021-power}.
VPT is especially effective in the low-data regime, and maintains its advantage across data scales.
Finally, VPT is competitive for a range of Transformer scales and designs (ViT-Base/Large/Huge, Swin).
Put together, our results suggest that VPT is one of the most effective ways of adapting ever-growing vision backbones.

\section{Related Work}\label{sec:related}

\noindent\textbf{Transformer}
models~\cite{vaswani2017attention} have gained huge success in NLP~\cite{devlin-etal-2019-bert,2020t5,brown2020gpt3}.
The triumph of the Transformer architecture also extends to various computer vision tasks, 
including image classification~\cite{dosovitskiy2020vit,liu2021swin}, object detection~\cite{carion2020end,li2021benchmarking}, semantic and panoptic segmentation~\cite{strudel2021segmenter,zheng2020rethinking,wang2021max}, video understanding~\cite{girdhar2019video,wang2022bevt,feichtenhofer2022masked} and few-shot learning~\cite{doersch2020crosstransformers}, surpassing previous state-of-the-art approaches.
Transformers are also being widely used in recent self-supervised pre-training methods~\cite{chen2021mocov3,he2021mae,bao2021beit}.
Given their superior performance and much larger scale compared to ConvNets, how to efficiently adapt Transformers to different vision tasks remains an important open problem. Our proposed VPT provides a promising path forward.

\noindent\textbf{Transfer learning}
has been extensively studied for vision tasks in the context of ConvNets~\cite{zhuang2020comprehensive} and many techniques have been introduced including side tuning~\cite{zhang2020side}, residual adapter~\cite{rebuffi2017learning}, bias tuning~\cite{cai2020tinytl}, \etc. 
Relatively little attention has been paid to vision Transformers adaptation and how well these aforementioned methods perform on this brand new type of architecture remains unknown. 
On the other hand, given the dominance of large-scale pre-trained Transformer-based Language Models (LM)~\cite{devlin-etal-2019-bert,2020t5,brown2020gpt3}, many approaches~\cite{he2022towards,guo2020parameter,hu2021lora} have been proposed to efficiently fine-tune LM for different downstream NLP tasks~\cite{wang2018glue,wang2019superglue}.
Among them, we focus on the following two representative methods in our experiments for benchmarking purposes: Adapters~\cite{pfeiffer2020AdapterHub} and BitFit~\cite{zaken2021bitfit}.

Adapters~\cite{houlsby2019parameter} insert extra lightweight modules inside each Transformer layer. One adapter module generally consists of a linear down-projection, followed by a nonlinear activation function, and a linear up-projection, together with a residual connection~\cite{pfeiffer2020adapterfusion,pfeiffer2020AdapterHub}.
Instead of inserting new modules, 
\cite{cai2020tinytl} proposed to update the bias term and freeze the rest of backbone parameters when fine-tuning ConvNets. BitFit~\cite{bao2021beit} applied this technique to Transformers and verified its effectiveness on LM tuning.
Our study demonstrates that VPT, in general, provides improved performance in adapting Transformer models for vision tasks, relative to the aforementioned two well-established methods in NLP.

\noindent\textbf{Prompting}~\cite{liu2021pre} originally refers to prepending language instruction to the input text so that a pre-trained LM can ``understand'' the task.
With manually chosen prompts, GPT-3 shows strong generalization to downstream transfer learning tasks even in the few-shot or zero-shot settings~\cite{brown2020gpt3}. In addition to the follow-up works on how to construct better prompting texts~\cite{shin2020autoprompt,jiang2020can}, recent works propose to treat the prompts as task-specific continuous vectors and directly optimize them via gradients during fine-tuning, namely Prompt Tuning~\cite{li-liang-2021-prefix,lester-etal-2021-power,liu2021p}. 
Compared to full fine-tuning, it achieves comparable performance but with 1000$\times$ less parameter storage. 
Although prompting has also been applied to vision-language models recently~\cite{radford2021learning,zhou2021learning,ju2021prompting,yao2021cpt,ge2022domain}, prompting is still limited to the input of \emph{text} encoders. 
Due to the disparity between vision and language modalities, in this paper we ask: can the same method can be applied successfully to image encoders?
We are the first work (see related concurrent works~\cite{sandler2022fine,wang2022learning,conder2022efficient,bahng2022visual}) to tackle this question and investigate the generality and feasibility of visual prompting via  \emph{extensive} experiments spanning multiple kinds of recognition tasks across multiple domains and backbone architectures.

\begin{figure}[t]
\centering
\includegraphics[width=\textwidth]{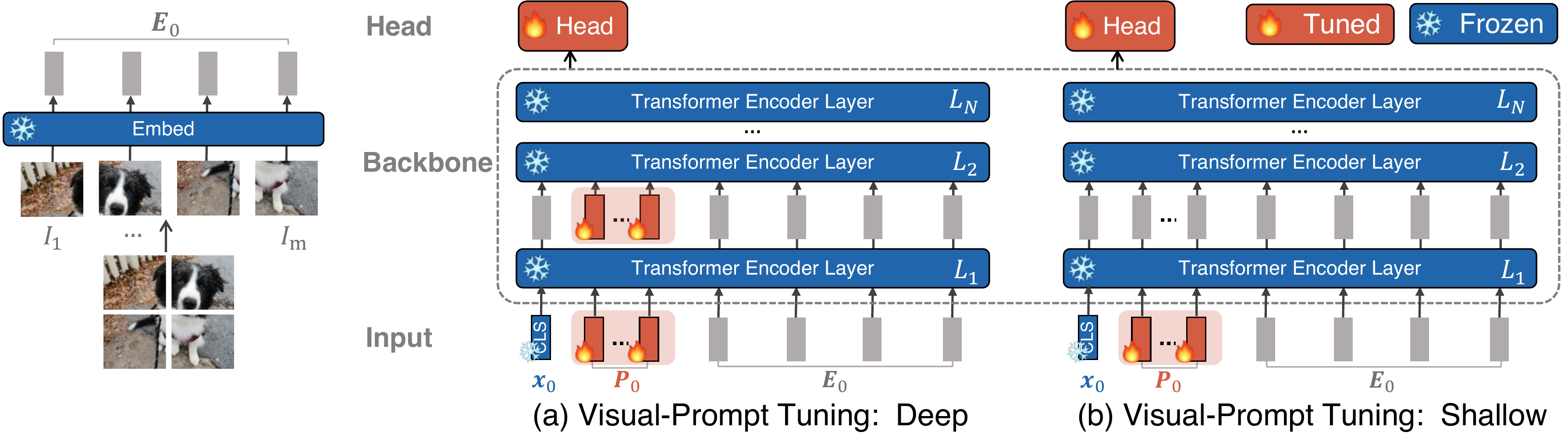}
\caption{Overview of our proposed \visualprompt{}. We explore two variants: (a) prepend a set of learnable parameters to each Transformer encoder layer's input (\deepprompt{}); (b) only insert the prompt parameters to the first layer's input (\shallowprompt{}).
During training on downstream tasks, only the parameters of prompts and linear head are updated while the whole Transformer encoder is frozen.
}
\label{fig:method}
\end{figure}

\section{Approach}\label{sec:method}

We propose Visual-Prompt Tuning (\vprompt{}) for adapting large pre-trained vision Transformer models. 
\vprompt{} injects a small number of learnable parameters into Transformer's input space and keeps the backbone frozen during the downstream training stage. 
The overall framework is presented in \cref{fig:method}.
We first define the notations in~\cref{subsec:method_pre}, then describe VPT formally in~\cref{subsec:method_vp}.

\subsection{Preliminaries}
\label{subsec:method_pre}

For a plain Vision Transformer (ViT)~\cite{dosovitskiy2020vit} with $N$ layers, an input image is divided into $m$ fixed-sized patches 
$\{I_j\in\R^{3\times h\times w}\mid j\in\N, 1\le j\le m\}$. $h, w$ are the height and width of the image patches.
Each patch is then first embedded into $d$-dimensional latent space with positional encoding:
\begin{align}
\label{eq:embed}
\vec{e}_0^j = \texttt{Embed}(I_j) &&\vec{e}_0^j\in\R^{d}, j = 1,2, \ldots m \eqdot
\end{align}

We denote the collection of image patch embeddings, 
$\vec{E}_{i}=\{\vec{e}_i^j\in\R^d\mid j\in\N, 1\le j\le m\}$,  
as inputs to the ($i$+$1$)-th Transformer layer ($L_{i+1}$). Together with an extra learnable classification token ($\texttt{[CLS]}$), the whole ViT is formulated as:
\begin{align}
\label{eq:vit}
    [\vec{x}_i, \vec{E}_i] &= L_i([\vec{x}_{i-1}, \vec{E}_{i-1}])  &&i=1, 2, \ldots, N \\
    \vec{y} &= \texttt{Head}(\vec{x}_N)\eqcomma
\end{align}
where $\vec{x}_{i}\in\R^{d}$ denote $\texttt{[CLS]}$'s embedding at $L_{i+1}$'s input space. 
$[\cdot,\cdot]$ indicates stacking and 
concatenation on the sequence length dimension, \ie, $[\vec{x}_{i}, \vec{E}_{i}]\in\R^{(1+m)\times d}$. 
Each layer $L_i$ consists of Multiheaded Self-Attention (MSA) and Feed-Forward Networks (FFN) together with LayerNorm~\cite{ba2016layer} and residual connections~\cite{he2016rn}. 
A neural classification head is used to map the final layer's $\texttt{[CLS]}$ embedding, $\vec{x}_{N}$, into a predicted class probability distribution $\vec{y}$.\footnote{Some Transformer architectures in Vision such as~\swin{}~\cite{liu2021swin} do not use $\texttt{[CLS]}$ and treat global pooled $\vec{E}_N$ as input for \texttt{Head}. We follow their designs when adapting VPT to these Transformer variants. See~\cref{supsec:detail} for more details.}

\subsection{Visual-Prompt Tuning (\vprompt{})}
\label{subsec:method_vp}

Given a pre-trained Transformer model, we introduce a set of $p$ continuous embeddings of dimension $d$, \ie{},~\emph{prompts}, in the input space after the \texttt{Embed} layer.
Only the task-specific prompts are being updated during fine-tuning, while the Transformer backbone is kept frozen. 
Depending on the number of Transformer layers involved, our approach has two variants, \shallowprompt{} and \deepprompt{}, as shown in~\cref{fig:method}.

\para{VPT-Shallow.}
Prompts are inserted into the first Transformer layer $L_1$ only.
Each prompt token is a learnable $d$-dimensional vector.
A collection of $p$ prompts is denoted as 
$\vec{P}=\{\vec{p}^k\in\R^d\mid k\in\N, 1\le k\le p\}$, the shallow-prompted ViT is: 
\begin{align}
\label{eq:shallow}
    [\vec{x}_{1}, \vec{Z}_{1}, \vec{E}_{1}] &= \textcolor{prompt_blue}{L_1}([\textcolor{prompt_blue}{\vec{x}_{0}}, \textcolor{prompt_red}{\vec{P}},\vec{E}_{0}])   \\
    [\vec{x}_{i}, \vec{Z}_{i}, \vec{E}_{i}] &= \textcolor{prompt_blue}{L_i}([\vec{x}_{i-1}, \vec{Z}_{i-1},\vec{E}_{i-1}])  &&i=2, 3, \ldots, N\\
    \vec{y} &= \textcolor{prompt_red}{\texttt{Head}}(\vec{x}_N)\eqcomma
\end{align}
where $\vec{Z}_{i}\in\mathbb{R}^{p\times d}$ represents the features computed by the $i$-th Transformer layer, and $[\vec{x}_{i}, \vec{Z}_{i}, \vec{E}_{i}]\in\R^{(1+p+m)\times d}$.
The colors \textcolor{prompt_red}{$\bullet$} and \textcolor{prompt_blue}{$\bullet$} indicate \textcolor{prompt_red}{learnable} and \textcolor{prompt_blue}{frozen} parameters, respectively.
Notably for ViT, $\vec{x}_N$ is invariant to the location of prompts since they are inserted after positional encoding, \eg, $[\vec{x}_{0}, \vec{P},\vec{E}_{0}]$ and $[\vec{x}_{0}, \vec{E}_{0}, \vec{P}]$ are mathematically equivalent. This also applies to VPT-Deep.

\para{VPT-Deep.}
Prompts are introduced at \emph{every} Transformer layer's input space. For ($i$+$1$)-th Layer 
$L_{i+1}$, we denote the collection of input learnable prompts as 
$\vec{P}_{i}=\{\vec{p}_i^k\in\R^d\mid k\in\N, 1\le k\le m\}$. The deep-prompted ViT is formulated as:
\begin{align}
\label{eq:deep}
    [\vec{x}_i, \underline{\hspace{0.3cm}}, \vec{E}_i] &= \textcolor{prompt_blue}{L_i}([\vec{x}_{i-1}, \textcolor{prompt_red}{\vec{P}_{i-1}},\vec{E}_{i-1}]) &&i=1, 2, \ldots, N\\
    \vec{y} &= \textcolor{prompt_red}{\texttt{Head}}(\vec{x}_N)\eqdot
\end{align}

\para{Storing Visual Prompts.} \vprompt{} is beneficial in presence of multiple downstream tasks.
We only need to store the learned prompts and classification head for each task and re-use the original copy of the pre-trained Transformer model, significantly reducing the storage cost.
For instance, given a \vit{}-Base with 86 million (M) parameters and $d=768$, 50 shallow prompts and deep prompts yield additional $p\times d=50\times 768=0.038$M, and $N\times p\times d=0.46$M parameters, amounting to only 0.04$\%$ and 0.53$\%$ of all \vit{}-Base parameters, respectively.

\section{Experiments}\label{sec:exp}

We evaluate VPT for a wide range of downstream recognition tasks with pre-trained Transformer backbones across scales.
We first describe our experimental setup in~\cref{subsec:evalsetup}, including the pre-trained backbone and downstream tasks, and a brief introduction of alternative transfer learning methods. Then we demonstrate the effectiveness and practical utility of our method in \cref{subsec:exp_results}.
We also systematically study how different design choices would affect performance (\cref{subsec:ablate}),
which leads to an improved understanding of our approach.

\subsection{Experiment Setup}
\label{subsec:evalsetup}

\noindent\textbf{Pre-trained Backbones.}
We experiment with two Transformer architectures in vision, Vision Transformers (\vit{})~\cite{dosovitskiy2020vit} and Swin Transformers (\swin{}~\cite{liu2021swin}). 
All backbones in this section are pre-trained on \imagenet{}-21k~\cite{imagenet_cvpr09}. We follow the original configurations, \eg, number of image patches divided, existence of \texttt{[CLS]}, \etc.
More details are included in~\cref{supsec:detail}.

\noindent\textbf{Baselines.}
We compare both variants of VPT with other commonly used fine-tuning protocols: 

\begin{enumerate}[nosep, label=(\alph*), font=\small\ttbf,] 
\item \fullft{}: fully update \emph{all} backbone and classification head parameters.

\item Methods that 
focus on the classification head. They treat the pre-trained backbone as a feature extractor, whose weights are fixed during tuning:
\begin{itemize}[leftmargin=0.0em, topsep=0.15mm]
    \item \linear{}: only use a linear layer as the classification head.
    \item \partialft{}-$k$: fine-tune the last $k$ layers of backbone while freezing the others, as adopted in~\cite{yosinski2014transferable,zhang2016colorful,noroozi2016unsupervised,he2021mae}. It redefines the boundary of backbone and classification head.
    \item \mlp{}-$k$: utilize a  multilayer perceptron (MLP) with $k$ layers, instead of a linear layer, as classification head.
\end{itemize}

\item 
Methods that update a subset backbone parameters or add new trainable parameters to backbone during fine-tuning:
\begin{itemize}[leftmargin=0.0em, topsep=0.15mm]
\item \sidetune{}~\cite{zhang2020side}: train a ``side'' network and linear interpolate between pre-trained features and side-tuned features before being fed into the head.
\item \bias{}~\cite{cai2020tinytl,zaken2021bitfit}: fine-tune only the bias terms of a pre-trained backbone.
\item \adapter{}~\cite{houlsby2019parameter,pfeiffer2020adapterfusion,pfeiffer2020AdapterHub}: insert new MLP modules with residual connection inside Transformer layers. 
\end{itemize}
\end{enumerate}

\noindent\textbf{Downstream Tasks.}
We experiment on the following two collections of datasets:

\textit{FGVC} consists of 5 benchmarked Fine-Grained Visual Classification tasks including 
CUB-200-2011~\cite{WahCUB_200_2011}, 
NABirds~\cite{van2015nabirds}, 
Oxford Flowers~\cite{nilsback2008automated}, 
Stanford Dogs~\cite{Khosla_FGVC2011dogs} and
Stanford Cars~\cite{gebru2017cars}. 
If a certain dataset only has \train{} and \test{} sets publicly available, we randomly split the training set into \train{} (90\%) and \val{} (10\%), and rely on \val{} to select hyperparameters.

\textit{\vtab{}}~\cite{zhai2019vtab} is a collection of 19 diverse visual classification tasks, which are organized into three groups: 
\textit{Natural} - tasks that contain natural images captured using standard cameras;
\textit{Specialized} - tasks that contain images captured via specialized equipment, such as medical and satellite imagery;
and \textit{Structured} - tasks that require geometric comprehension like object counting.
Each task of VTAB contains 1000 training examples. Following~\cite{zhai2019vtab}, we use the provided 800-200 split of the \train{} set to determine hyperparameters and run the final evaluation using the full training data. We report the average accuracy score on \test{} set within three runs.

We report the average accuracy on the FGVC datasets, and the average accuracy on each of the three groups in VTAB.
The individual results on each task are in~\cref{subsec:results_supp}, as are image examples of these aforementioned tasks.

\setlength{\tabcolsep}{4pt}
\begin{table}[t]
\caption{\vit{}-B/16 pre-trained on supervised \imagenet-21k. 
For each method and each downstream task group,
we report the average test accuracy score and \texttt{number of wins in ($\cdot$)} compared to \fullft{}.
``Total params'' denotes total parameters needed for all 24 downstream tasks. 
``Scope'' denotes the tuning scope of each method.
``Extra params'' denotes the presence of additional parameters besides the pre-trained backbone and linear head.
Best results among all methods except \fullft{} are \textbf{bolded}.
\vprompt{} outshines the full fine-tuning 20 out of 24 cases with significantly less trainable parameters
}
\label{table:main_vitb}
\resizebox{\textwidth}{!}{
\begin{tabular}{
ll  !{\color{tabvline}\vrule}
r   !{\color{tabvline}\vrule}
cc  !{\color{tabvline}\vrule}
c !{\color{tabvline}\vrule}
rrrr }
\toprule
&\textbf{\vit{}-B/16 }
&\bf{Total}
&\multicolumn{2}{c!{\color{tabvline}\vrule}}{\bf{Scope}}
&\bf{Extra}
&\multirow{2}{*}{\bf{FGVC}}
&\multicolumn{3}{c}{\bf{\vtab{}}}
\\
&\bf{(85.8M)}
&\bf{params}
&\bf{Input} &\bf{Backbone} &\bf{params}
&&\bf{\scriptsize{Natural}} &\bf{\scriptsize{Specialized}} &\bf{\scriptsize{Structured}}
\\
\midrule
&Total \# of tasks &&&&
&5 &7 &4 &8
\\
\midrule
\band \ttbf{(a)}&\fullft{} & 24.02$\times$ & &\checkmark &
&88.54 &75.88 &83.36 &47.64
\\
\midrule
\multirow{3}{*}{\ttbf{(b)}}
&\linear{} &1.02$\times$ & & &
&79.32 (0)   &68.93 (1) &77.16 (1) &26.84 (0)
\\
&\partialft{}-1 &3.00$\times$ & & &
&82.63 (0) &69.44 (2) &78.53 (0) &34.17 (0)
\\
 &\mlp-3 &1.35$\times$ && &\checkmark
&79.80 (0) &67.80 (2) &72.83 (0) &30.62 (0)
\\

\midrule

\multirow{3}{*}{\ttbf{(c)}}&\sidetune & 3.69$\times$ &&\checkmark &\checkmark
&78.35 (0) &58.21 (0) &68.12 (0) &23.41 (0)
\\
&\bias{} &1.05$\times$ &&\checkmark &
&88.41 (3) &73.30 (3) &78.25 (0) &44.09 (2)
\\
&\adapter{} &1.23$\times$ &&\checkmark &\checkmark
&85.66 (2) &70.39 (4) &77.11 (0) &33.43 (0)
\\
\midrule
\multirow{2}{*}{\ttbf{(ours)}}
&\shallowprompt{} & 1.04$\times$ &\multirow{2}{*}{\checkmark} & &\multirow{2}{*}{\checkmark}
&84.62 (1) &76.81 (4) &79.66 (0) &46.98 (4)
\\
&\deepprompt{} &1.18$\times$ & &&
&\textbf{89.11} (\textbf{4})  

&\textbf{78.48} (\textbf{6}) &\textbf{82.43} (\textbf{2}) &\textbf{54.98} (\textbf{8})
\\
\bottomrule

\end{tabular}
}
\end{table}
\setlength{\tabcolsep}{1.4pt}

\subsection{Main Results}
\label{subsec:exp_results}

\cref{table:main_vitb} presents the results of fine-tuning a pre-trained \vit{}-B/16 on 
averaged across 4 diverse downstream task groups, comparing VPT to the other 7 tuning protocols. 
We can see that:

\begin{enumerate}[nosep, leftmargin=5mm]
\item \textbf{VPT-Deep outperforms \fullft{} (\cref{table:main_vitb}\texttt{(a)}) on 3 out of the 4 problem classes} (20 out of 24 tasks), while using significantly fewer total model parameters (1.18$\times$~\vs~24.02$\times$).
Thus, \emph{even if storage is not a concern},  \vprompt{} is a promising approach for adapting larger Transformers in vision. 
Note that this result is in contrast to comparable studies in NLP, where prompt tuning matches, but \emph{does not exceed} full fine-tuning~\cite{lester-etal-2021-power}.

\item \textbf{VPT-Deep outperforms all the other parameter-efficient tuning protocols (\cref{table:main_vitb}\texttt{(b},\texttt{c)}) across all task groups}, indicating that \deepprompt{} is the best fine-tuning strategy in storage-constrained environments.

\item Although sub-optimal than \deepprompt{}, \shallowprompt{} still offers non-trivial performance gain than head-oriented tuning methods in \cref{table:main_vitb}\texttt{(b)}, indicating that \shallowprompt{} is a worthwhile choice in deploying multi-task fine-tuned models if the storage constraint is severe.
\end{enumerate}

\begin{figure}[t]
\centering
\includegraphics[width=\textwidth]{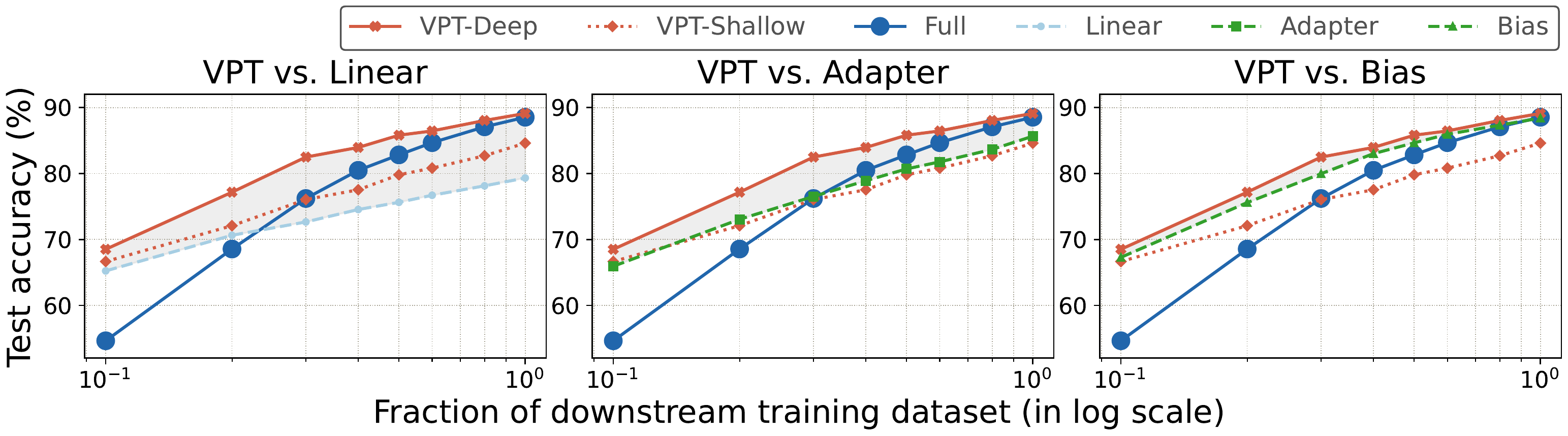}
\caption{
Performance comparison on different downstream data scales, averaged across 5 FGVC tasks.
\deepprompt{} is compared with \linear{} (left), \adapter{} (middle) and \bias{} (right).
Highlighted region shows the accuracy difference between \deepprompt{} and the compared method.
Results of \shallowprompt{} are \fullft{} presented in all plots for easy reference. 
The size of markers are proportional to the percentage of tunable parameters in log scale
}
\label{fig:main_fgvcsize}
\end{figure}

\para{VPT on different downstream data size.}
\label{para:exp_datasize}
We look at the impact of training data size on accuracy in the FGVC tasks (VTAB has only 1k training examples).
We vary the training data between 10\% and 80\% and compare all methods.
The same pre-trained \vit{}-B is used for downstream training. 
Task-averaged results for each method on different training data scales are presented in \cref{fig:main_fgvcsize}.

\cref{fig:main_fgvcsize} shows that \deepprompt{} outperforms all the other baselines across data scales.
Digging deeper, methods that use less trainable parameters, \ie, \vprompt{}, \linear{}, \adapter{}, \bias{}, dominate over \fullft{} in the low-data regimes.
This trend, however, is \emph{reversed} when more training data is available for \linear{} and \adapter{}. 
In contrast, \deepprompt{} still consistently outperforms \fullft across training data sizes.
Although \bias{} offers similar advantages, it still marginally under-performs \deepprompt{} across the board (\cref{fig:main_fgvcsize} right).

\begin{figure}[t]
\centering
\includegraphics[width=\textwidth]{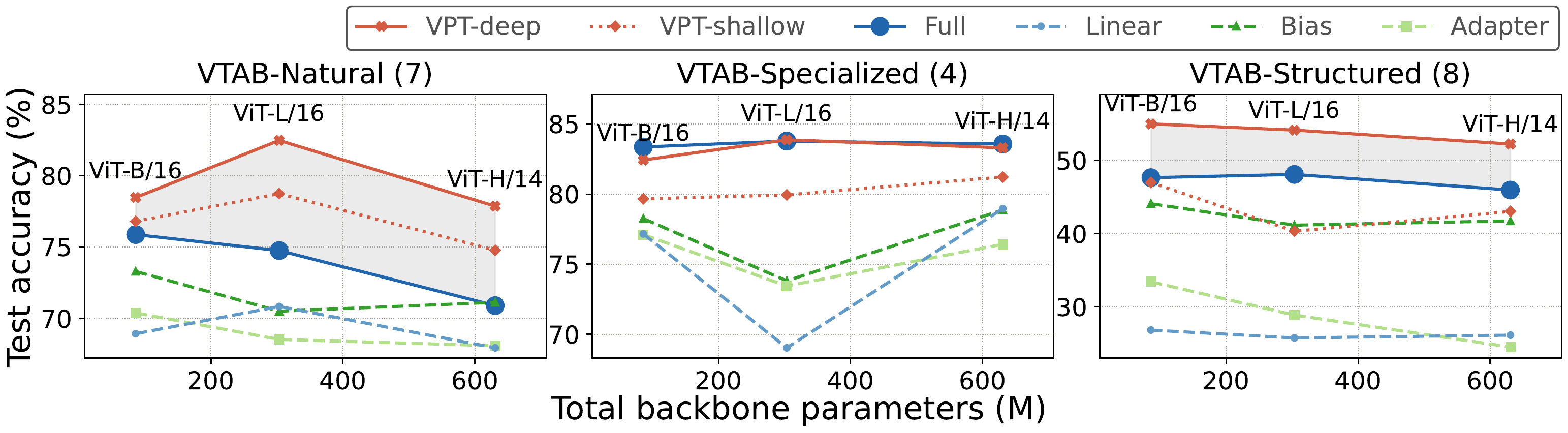}
\caption{\vprompt{}~\vs{}~\fullft{} across model scales (\vit{}-B, \vit{}-L and \vit{}-H), for 3 VTAB task groups.
Highlighted region shows the accuracy difference between \deepprompt{} and the full fine-tuning (\fullft{}).
The size of markers are proportional to the percentage of trainable parameters in log scale
}
\label{fig:modelsize}
\end{figure}

\para{VPT on different backbone scales.}
\cref{fig:modelsize} shows \vtab{} performance under 3 different backbone scales: ViT-\textbf{B}ase/\textbf{L}arge/\textbf{H}uge.
\deepprompt{} is significantly better than \linear{} and \shallowprompt{} across all 3 backbone choices and 3 subgroups of \vtab{}.
More importantly, the advantages of \deepprompt{} over \fullft{} indeed still hold as the model scale increases, \ie, \deepprompt{} significantly outperforms \fullft{} on \textit{Natural} and \textit{Structured} groups, while offering nearly equivalent performance on \textit{Specialized}.

\setlength{\tabcolsep}{4pt}
\begin{table}[t]
\begin{center}
\caption{Different Transformer architecture: \swin{}-B pre-trained on supervised \imagenet-21k as backbone. 
For each method and each downstream task group,
we report the average test accuracy score and \texttt{number of wins in ($\cdot$)} compared to \fullft{}.
The column ``Total params'' denotes total parameters needed for all 19 downstream tasks.
Best results among all methods except \fullft{} are \textbf{bolded}
}
\label{table:main_swinb}
\resizebox{0.6\textwidth}{!}{
\begin{tabular}{ll r rrr}
\toprule
&\textbf{\swin{}-B } &\bf{Total}
&\multicolumn{3}{c}{\bf{\vtab{}}}
\\
&\bf{(86.7M)}&\bf{params}
&\bf{Natural}
&\bf{Specialized}
&\bf{Structured}
\\
\midrule
&Total \# of tasks&
 &7 &4 &8
\\
\midrule
\band \ttbf{(a)}&\fullft{} & 19.01$\times$  
&79.10 &86.21 &59.65
\\
\midrule
\multirow{3}{*}{\ttbf{(b)}}&\linear{} &1.01$\times$ 
&73.52 (5) &80.77 (0) &33.52 (0)
\\
&\mlp{}-3 &1.47$\times$  
&73.56 (5) &75.21 (0) &35.69 (0)
\\
&\partialft{} &3.77$\times$  
&73.11 (4) &81.70 (0) &34.96 (0)
\\
\midrule
\ttbf{(c)}&\bias{} &1.06$\times$ 
&74.19 (2) &80.14 (0) &42.42 (0)
\\
\midrule
\multirow{2}{*}{\ttbf{(ours)}}
&\shallowprompt{} & 1.01$\times$ 
 &\textbf{79.85} (\textbf{6}) &82.45 (0) &37.75 (0)
\\
&\deepprompt{} &1.05$\times$ 
&76.78 (\textbf{6}) &\textbf{84.53} (0) &\textbf{53.35} (0)
\\
\bottomrule
\end{tabular}
}
\end{center}
\end{table}
\setlength{\tabcolsep}{1.4pt}

\para{VPT on hierarchical Transformers.}
We extend \vprompt{} to Swin~\cite{liu2021swin}, which employs MSA within local shifted windows and merges patch embeddings at deeper layers. For simplicity and without loss of generality, we implement VPT in the most straightforward manner:
the prompts are attended within the local windows, but are ignored during patch merging stages.
The experiments are conducted on the~\imagenet{}-21k supervised pre-trained Swin-\textbf{B}ase. 
\vprompt{} continues to outperform other parameter-efficient fine-tuning methods (\texttt{b, c}) for all three subgroups of VTAB \cref{table:main_swinb}, though in this case \fullft{} yields the highest accuracy scores overall (at a heavy cost in total parameters). 

It is surprising that the advantage of \deepprompt{} over \shallowprompt{} diminishes for \textit{Natural}: \shallowprompt{} yields slightly better accuracy scores than full fine-tuning.

\subsection{Ablation on Model Design Variants}
\label{subsec:ablate}
We ablate different model design choices on the supervised ImageNet-21k pre-trained ViT-Base and evaluate them on VTAB, with same setup in \cref{table:main_vitb}. See more in~\cref{supp_analysis}.

\begin{figure}[t]
\centering
\includegraphics[width=0.95\textwidth]{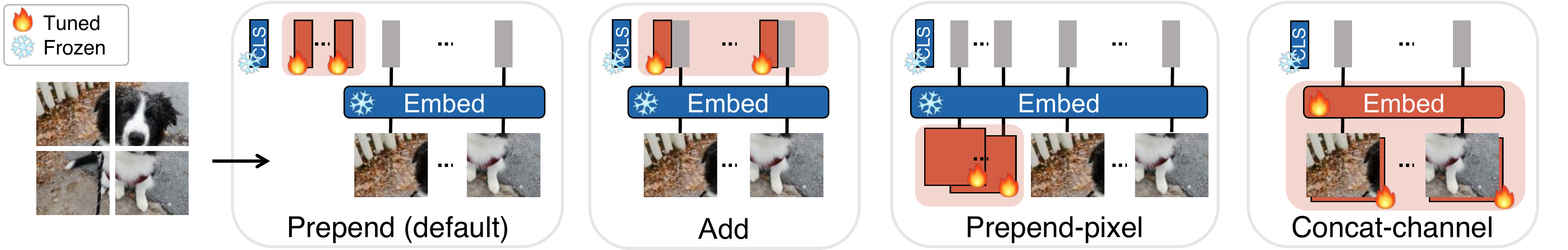}
\includegraphics[width=\textwidth]{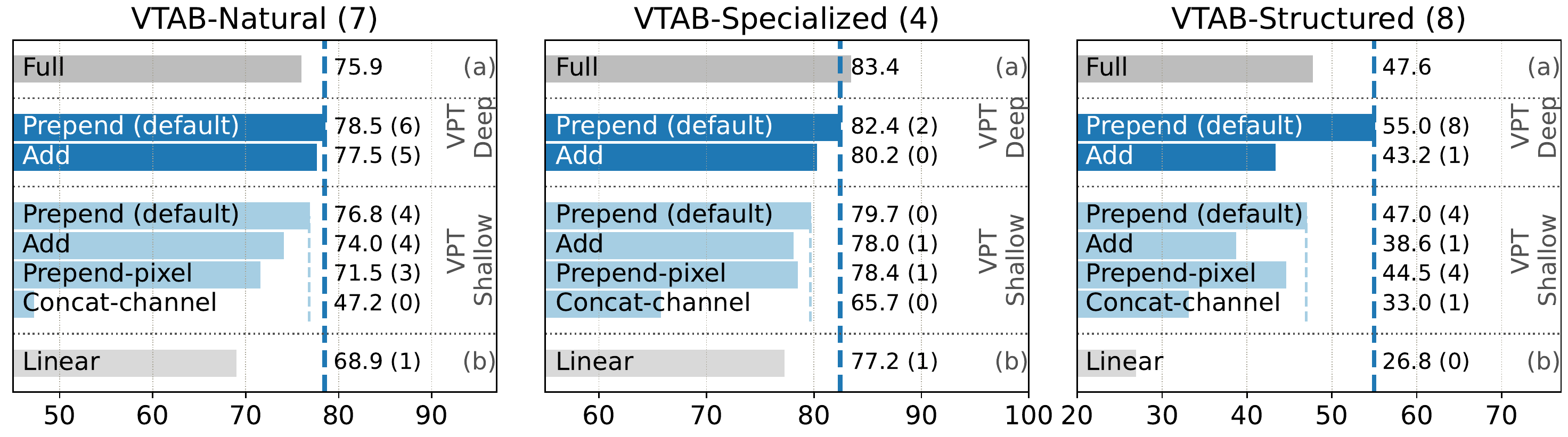}
\caption{Ablation on prompt location. We illustrate different location choices at top, and present the results at bottom.
For easy comparison, two blue dashed lines represent the performance of the default \deepprompt{} and \shallowprompt{} respectively
}
\label{fig:ablate_loc}
\end{figure}

\para{Prompt Location.}
An important distinction between \vprompt{} and other methods is the extra learnable parameters introduced as \textit{inputs} for the Transformer layers. 
\cref{fig:ablate_loc} ablates different choices on how and where to insert prompts in the input space, and how they would affect the final performance.

\textit{Prepend or Add?}
Instead of prepending prompts to the sequence of the image patches embeddings $\vec{E}_i$ as described in~\cref{subsec:method_vp}, 
another option is to directly \emph{add} prompts element-wise to those embeddings, keeping the Transformer's input sequence length the same as before. 
Though this variant is competitive to \fullft{} in some cases (\eg, VTAB-\textit{Natural}), its performance generally falls behind with the default \texttt{Prepend} in both deep and shallow settings.
More discussion on this phenomenon is in~\cref{app:seq}.

\textit{Latent or pixel space?} Instead of inserting the prompts as latent vectors for the first Transformer layer, one could introduce prompts in the \textit{pixel} level before the \texttt{Embed} layer in~\cref{eq:embed},
\ie, \texttt{Prepend-pixel} and \texttt{Concat-channel}.
\cref{fig:ablate_loc} shows that the adaption performance \emph{decreases} for these two variants.
For example, the accuracy score of prepending shallow prompts before the projection layer (\texttt{Prepend-pixel}) drops 6.9$\%$, compared to the default prepending in the embedding space (\texttt{Prepend}) on VTAB-\textit{Natural}. The performance further deteriorates (even as large as 30 accuracy scores drop on VTAB-\textit{Natural}) if we instead concatenate a new channel to the input image (\texttt{Concat-channel}). These observations suggest that it's easier for prompts to learn condensed task-dependent signals in the latent input space of Transformers.

\begin{figure}[t]
\centering
\includegraphics[width=0.95\textwidth]{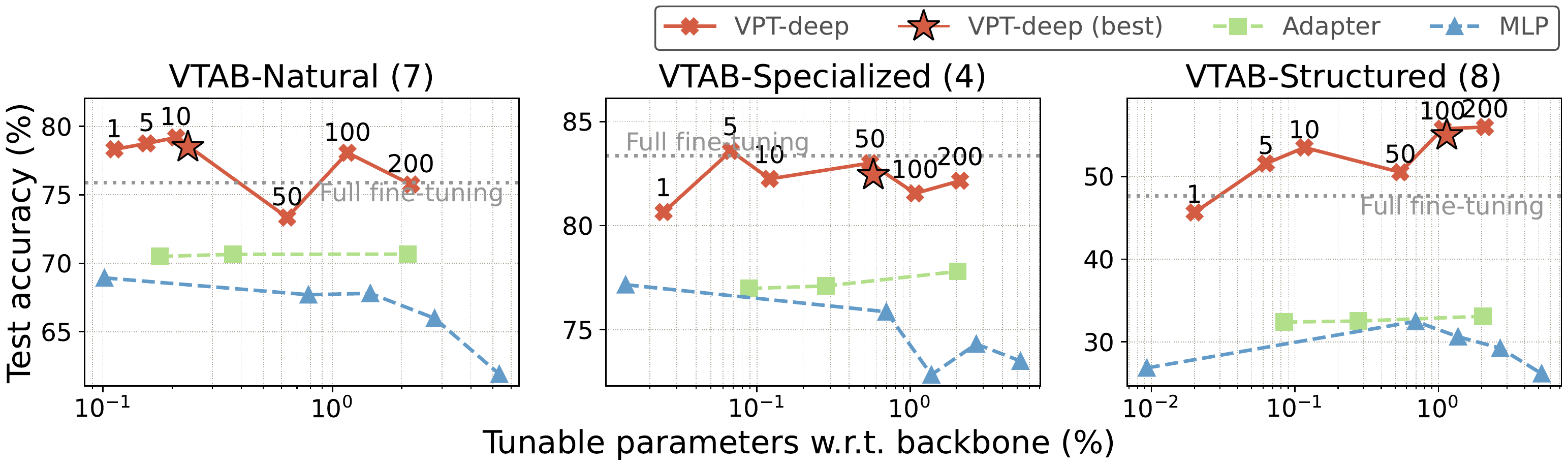}
\caption{Ablation on prompt length. We vary the number of prompts for \deepprompt{} and show the averaged results for each VTAB subgroup. 
The averaged best \deepprompt{} results for each task is also shown for easy reference
}
\label{fig:ablate_length}
\end{figure}

\para{Prompt Length.} This is the only additional hyper-parameter needed to tune for VPT compared to full fine-tuning. 
For easy reference, we also ablate two other baselines on their individual additional hyper-parameters, \ie, number of layers for \mlp{} and reduction rate for \adapter{}. As shown in \cref{fig:ablate_length}, the optimal prompt length varies across tasks.
Notably, even with as few as only \emph{one} prompt, \deepprompt{} still significantly outperforms the other 2 baselines, and remains competitive or even better compared to full fine-tuning on VTAB-\textit{Structured} and \textit{Natural}.

\begin{figure}[t]
\centering
\includegraphics[width=0.95\textwidth]{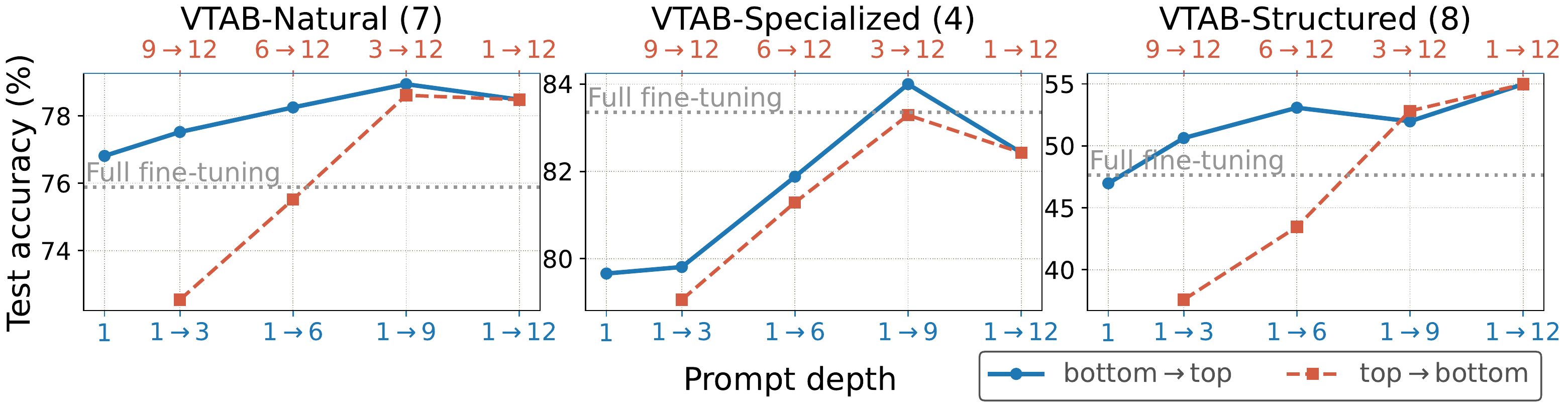}
\caption{Ablation on prompt depth.
We select the best prompt length for each variant with \val{} sets.
$i\rightarrow j$ indicates the Transformer layer indices that prompts are inserted into. The 1-st layer refers to the one closest to input. ViT-B has 12 layers in total
}
\label{fig:ablate_depth}
\end{figure}

\subsubsection{Prompt Depth.}
\cref{fig:ablate_depth} ablates which and how many layers to insert prompts.
Each variant reports the best prompt length selected with \val{} set.
\vprompt{}'s performance is positively correlated with the prompt depth in general. Yet the accuracy drops if we insert prompts from \textcolor{prompt_red}{top to bottom}, suggesting that prompts at earlier Transformer layers matter more than those at later layers.

\begin{figure}[t]
\centering
\includegraphics[width=0.95\textwidth]{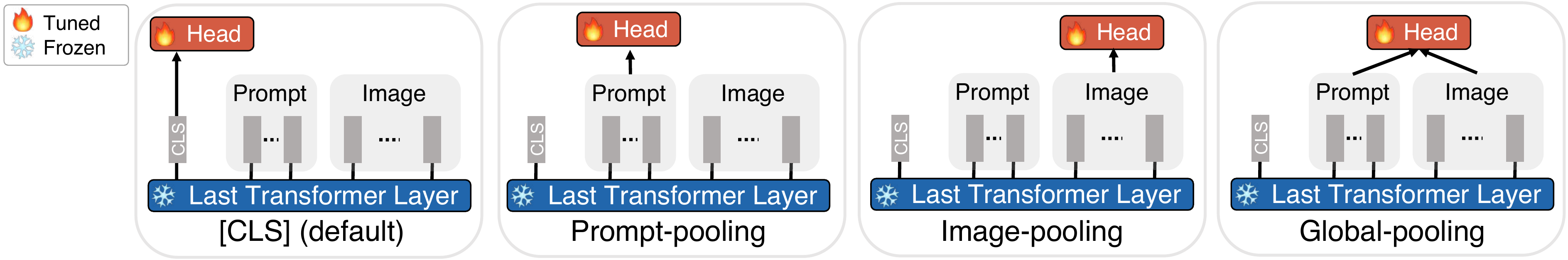}
\includegraphics[width=\textwidth]{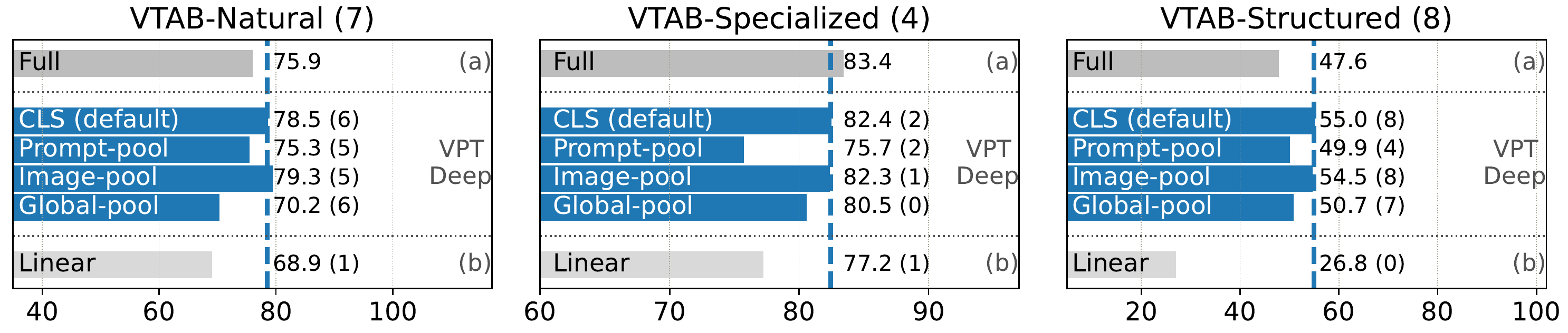}
\caption{
Ablation on final output. Illustration of different strategies is included at top, and results of those are presented at the bottom section.
For easy comparison, the blue dashed line represents the performance of default \deepprompt{}
}
\label{fig:ablate_output}
\end{figure}

\para{Final Output.}
Following the original configuration of \vit{}, we use 
the final embedding of $\texttt{[CLS]}$, \ie, $\vec{x}_N$, as the classification head input, which is also the default setting in our \vit{} experiments. 
As shown in \cref{fig:ablate_output}, 
if we use the average pooling on image patch output embeddings 
$\vec{E}_N$ as final output (\texttt{Image-pool}), the results essentially remain the same (\eg, 82.4 \vs~82.3 for VTAB-\textit{Specialized}).
However, if the pooling involves final prompt outputs $\vec{Z}_N$ (\texttt{Prompt-pool} and \texttt{Global-pool}), the accuracy could drop as large as 8 points.

\begin{figure}[t]
\centering
\includegraphics[width=0.95\textwidth]{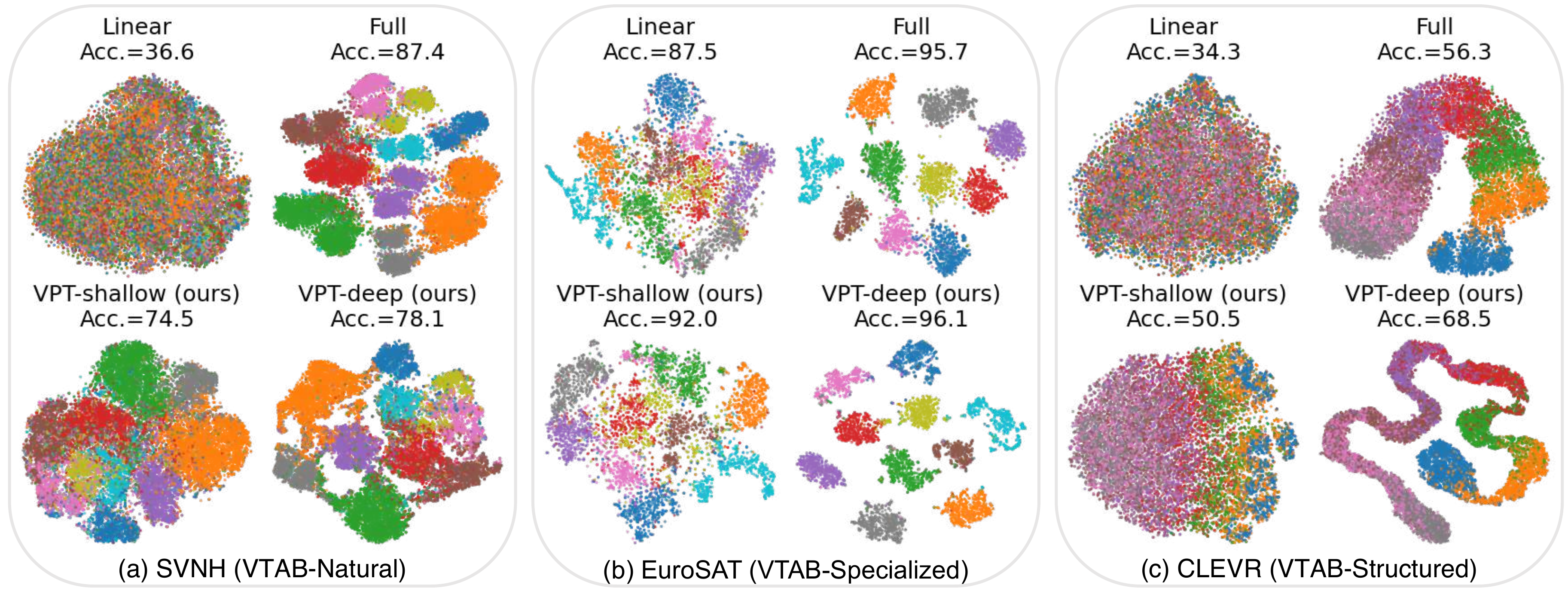}
\caption{t-SNE visualizations of the final \texttt{[CLS]} embedding \textbf{$\vec{x}_N$} of 3 VTAB tasks from the \test{} set, from~\cref{table:main_vitb}.
\vprompt{} could produce linearly separable features without updating backbone parameters
}
\label{fig:tsne}
\end{figure}

\section{Analysis and Discussion}
\label{sec:ana}

\para{Visualization.}
\cref{fig:tsne} shows t-SNE~\cite{van2008visualizing} visualizations of $\mathbf{x_N}$, \ie, embeddings of \texttt{[CLS]} after the last Transformer layer and before the classification head, for 3 tasks in VTAB (SVNH~\cite{netzer2011reading}, EuroSAT~\cite{helber2019eurosat}, Clevr/count~\cite{johnson2017clevr}), one for each subgroup. 
All plots show that \deepprompt{} enables linearly separable representations while using less parameters than \fullft{}.
We also observe that extra tunable parameters for every Transformer layer (\deepprompt{}) improve the performance, compared to \shallowprompt{}, which only inserts prompts for the first layer's input.
Interestingly on Clevr/count (\cref{fig:tsne}(c)), \deepprompt{} and \fullft{} recover the underlying manifold structure of the task (counting objects in images~\vs{} street number or landscape recognition), unlike \shallowprompt{} and \linear{}.

\setlength{\tabcolsep}{4pt}
\begin{table}
\scriptsize
\begin{center}
\caption{
Semantic Segmentation: ADE20k~\cite{zhou2019ade20} validation results with SETR~\cite{zheng2020rethinking} on ViT-L.
The best mIoU scores among all methods but \fullft{} are \textbf{bolded}. 
Results of fully fine-tuning a \rn{}-101~\cite{chen2018encoder} are included.
SS/MS: single/multi-scale inference
}
\label{table:main_seg}
\resizebox{0.8\textwidth}{!}{
\begin{tabular}{l r
!{\color{tabvline}\vrule}
rrrr !{\color{tabvline}\vrule}
c
}
\toprule
\textbf{Backbone} &
\multicolumn{5}{c}{ViT-L/16}
&\rn{}-101
\\
\cmidrule{2-7}
\textbf{Method}&\fullft{}~\cite{zheng2020rethinking} &\textsc{Head Only} & \bias{} &\deepprompt{} &\promptbias{}
&\fullft{}~\cite{chen2018encoder}
\\
\midrule
\textbf{mIoU}-SS
&48.31
&35.12  &43.40  &42.11  &\textbf{44.04}
&45.47
\\
\textbf{mIoU}-MS
&50.07
&37.46  &45.33  &44.06  &\textbf{45.63}
&46.27
\\
\textbf{Tunable params (M)}
&318.31 &13.18 &13.46 &13.43 &15.79 &63.0
\\
\bottomrule
\end{tabular}
}
\end{center}
\end{table}
\setlength{\tabcolsep}{1.4pt}

\para{Apply VPT to more vision tasks.}
We explore the feasibility of \vprompt{} beyond visual classification, by evaluating ADE20K~\cite{zhou2019ade20} semantic segmentation task with a Transformer model, SETR-PUP~\cite{zheng2020rethinking}. It adds a standard ConvNet head to the ViT backbone to perform segmentation. The de-facto approach is still fully fine-tuning the pre-trained backbone together with the ConvNet head (\fullft{}). We include two more protocols for comparison: only update the head layers (\textsc{Head Only}), update head layers and bias vectors in the backbone (\bias{}). 
In~\cref{table:main_seg}, we report \val{} mIoU results with and without multi-scale inference.
Though parameter-efficient protocols could not compete with \fullft{}, \vprompt{} is still comparable with \bias{}.
Notably, \vprompt{} offers competitive results to
a fully fine-tuned state-of-the-art ConvNet model (DeepLab v3+~\cite{chen2018encoder}), while tuning significantly less parameters (15M \vs{} 64M, respectively).

\setlength{\tabcolsep}{4pt}
\begin{table}[t]
\caption{Different pre-trained objectives: \mae{}~\cite{he2021mae} and \moco{}~\cite{chen2021mocov3} with a \vit{}-B backbone.
For each method and each downstream task group,
we report the average test accuracy score and \texttt{number of wins in ($\cdot$)} compared to \fullft{}.
``Total params'' denotes total parameters needed for all 24 downstream tasks. 
Best results among all methods except \fullft{} are \textbf{bolded}
}
\label{table:main_ssl}
\begin{center}
\resizebox{0.8\textwidth}{!}{
\begin{tabular}{
ll 
r rrr  !{\color{tabvline}\vrule}
r rrr 
}
\toprule
&&\multicolumn{4}{c}{\textbf{\mae{}}}
&\multicolumn{4}{c}{\textbf{\moco{}}}
\\
\cmidrule{3-10}
&\textbf{\vit{}-B/16 }
&\bf{Total} &\multicolumn{3}{c!{\color{tabvline}\vrule}}{\bf{\vtab{}}}
&\bf{Total} &\multicolumn{3}{c}{\bf{\vtab{}}}

\\
&\bf{(85.8M)} &\bf{params} &\bf{\scriptsize{Natural}} &\bf{\scriptsize{Specialized}} &\bf{\scriptsize{Structured}}
&\bf{params} &\bf{\scriptsize{Natural}} &\bf{\scriptsize{Specialized}} &\bf{\scriptsize{Structured}}
\\
\midrule
&Total \# of tasks
&&7 &4 &8
&&7 &4 &8
\\
\midrule
\band \ttbf{(a)}&\fullft{} 
&19.01$\times$ &59.29 &79.68 &53.82
&19.01$\times$ &71.95 &84.72 &51.98
\\
\midrule
\multirow{2}{*}{\ttbf{(b)}}&\linear{} 
&1.01$\times$ &18.87 (0) &53.72 (0) &23.70 (0)
&1.01$\times$ &67.46 (4) &81.08 (0) &30.33 (0)
\\
&\partialft{}-1
&2.58$\times$ &\textbf{58.44} (\textbf{5}) &\textbf{78.28} (1) &47.64 (\textbf{1})
&2.58$\times$ &72.31 (\textbf{5}) &\textbf{84.58} (\textbf{2}) &47.89 (1)
\\
\midrule
\multirow{2}{*}{\ttbf{(c)}}
&\bias{}
&1.03$\times$ &54.55 (1) &75.68 (1) &\textbf{47.70} (0)
&1.03$\times$ &72.89 (3) &81.14 (0) &\textbf{53.43} (\textbf{4})
\\
&\adapter{}
&1.17$\times$ &54.90 (3) &75.19 (1) &38.98 (0)
&1.22$\times$ &\textbf{74.19} (4) &82.66 (1) &47.69 (2)
\\
\midrule
\multirow{2}{*}{\ttbf{(ours)}}
&\shallowprompt{} 
&1.01$\times$ &39.96 (1)  &69.65 (0)  &27.50 (0)
&1.01$\times$ &67.34 (3) &82.26 (0) &37.55 (0)
\\
&\deepprompt{}
&1.04$\times$ &36.02 (0)  &60.61 (1)  &26.57 (0)
&1.01$\times$ &70.27 (4) &83.04 (0) &42.38 (0)
\\
\bottomrule
\end{tabular}
}
\end{center}
\end{table}
\setlength{\tabcolsep}{1.4pt}

\para{Apply VPT to more pre-training methods.}
In addition to the backbones pre-trained with labeled data, we experiment with two self-supervised objectives: \mae{}~\cite{he2021mae} and \moco{}~\cite{chen2021mocov3}.
\cref{table:main_ssl} reports the results on \vtab{} with \vit{}-B.
We observe that both variants of \vprompt{} surpass \linear{}, yet the comparisons among other techniques are less conclusive.
For \mae{}, other parameter-efficient methods, \eg, \partialft{}-1, outperform both \vprompt{} and \linear{}.
In the case of \moco{}, \vprompt{} no longer holds the best performance, though it is still competitive with the others.
This suggests that these two self-supervised \vit{}s are fundamentally different from the supervised ones in previous sections. Exactly why and how these differences arise remain open questions.

\setlength{\tabcolsep}{4pt}
\begin{table}[t]
\caption{
Apply VPT to ConvNets: \rn{}-50 and \rnx{}-Base. 
For each method and each downstream task group,
we report the average test accuracy score and \texttt{number of wins in ($\cdot$)} compared to \fullft{}.
``Total params'' denotes total parameters needed for all 19 downstream tasks.
Best results among all methods except \fullft{} are \textbf{bolded}
}
\label{table:main_rnxrn}
\begin{center}
\resizebox{0.875\textwidth}{!}{
\begin{tabular}{
ll 
r rrr  !{\color{tabvline}\vrule}
r rrr 
}
\toprule
&&\multicolumn{4}{c}{\textbf{\rnx{}-Base (87.6M)}}
&\multicolumn{4}{c}{\textbf{\rn{}-50 (23.5M)}}
\\
\cmidrule{3-10}
&
&\bf{Total} &\multicolumn{3}{c!{\color{tabvline}\vrule}}{\bf{\vtab{}}}
&\bf{Total} &\multicolumn{3}{c}{\bf{\vtab{}}}

\\
& &\bf{params} &\bf{\scriptsize{Natural}} &\bf{\scriptsize{Specialized}} &\bf{\scriptsize{Structured}}
&\bf{params} &\bf{\scriptsize{Natural}} &\bf{\scriptsize{Specialized}} &\bf{\scriptsize{Structured}}
\\
\midrule
&Total \# of tasks
&&7 &4 &8
&&7 &4 &8
\\
\midrule
\band \ttbf{(a)}&\fullft{} 
& 19.01$\times$ &77.97  &83.71  &60.41 
& 19.08$\times$  &59.72 &76.66  &54.08
\\
\midrule
\multirow{3}{*}{\ttbf{(b)}}&\linear{} 
&1.01$\times$ &74.48 (5) &81.50 (0) &34.76 (\textbf{1})
&1.08$\times$  &63.75 (\textbf{6}) &77.60 (\textbf{3}) &30.96 (0)
\\
&\partialft{}-1
&2.84$\times$  &73.76 (4) &81.64 (0) &39.55 (0)
&4.69$\times$  &64.34 (\textbf{6}) &\textbf{78.64} (2) &\textbf{45.78} (\textbf{1})
\\
&\mlp{}-3
&1.47$\times$ &73.78 (5) &81.36 (\textbf{1}) &35.68 (\textbf{1})
&7.87$\times$ &61.79 (\textbf{6}) &70.77 (1) &33.97 (0)
\\
\midrule
\ttbf{(c)}&\bias{}
&1.04$\times$ &69.07 (2) &72.81 (0) &25.29 (0)
&1.10$\times$ &63.51 (\textbf{6}) &77.22 (2) &33.39 (0)
\\
\midrule
\ttbf{(ours)}&\visualprompt{} 
& 1.02$\times$  &\textbf{78.48} (\textbf{6}) &\textbf{83.00} (\textbf{1}) &\textbf{44.64} (\textbf{1})
& 1.09$\times$  &\textbf{66.25} (\textbf{6}) &77.32 (2) &37.52 (0)
\\
\bottomrule
\end{tabular}
}
\end{center}
\end{table}
\setlength{\tabcolsep}{1.4pt}

\para{Apply VPT to ConvNets.}
We examine the idea of adding trainable parameters in the input space of ConvNets: padding both height and width by $p$ learnable prompt pixels for the input image. 
Though this operation seems unconventional, 
we implement VPT this way given there is no obvious solution to add location-invariant prompts similar to the Transformer counterparts.
In fact this approach has been explored before in the adversarial attack literature~\cite{elsayed2018adversarial}.
The value of $p$ in our experiment is 2 orders of magnitude smaller than previous work: \eg, 5 \vs 263. Most importantly, we cast this idea in the lens of transfer learning. See~\cref{vpt_ar} for more discussion.

\cref{table:main_rnxrn} presents the results for \rnx{}-B~\cite{liu2022convnet} (pre-trained on \imagenet{}-21k) and \rn{}-50~\cite{he2016rn} (pre-trained on \imagenet{}-1k), respectively.
\vprompt{} works well in a larger ConvNet backbone, \rnx{}-B, offering accuracy gains over other sparse tuning protocols (\texttt{b}, \texttt{c}), and outperforming \fullft{} on 8 out of 19 cases.
The advantages of \vprompt{}, however, diminish with smaller ConvNet (\rn{}-50), as there is no clear winner for all 19 \vtab{} tasks.

\section{Conclusion}

We present Visual Prompt Tuning, a new parameter-efficient approach to leverage large vision Transformer models for a wide range of downstream tasks.
VPT introduces task-specific learnable prompts in the input space, keeping the pre-trained backbone fixed.
We show that VPT can surpass other fine-tuning protocols (often including full fine-tuning) while dramatically reducing the storage cost.
Our experiments also raise intriguing questions
on fine-tuning dynamics of vision Transformers with different pre-training objectives, and how to transfer to broader vision recognition tasks in an efficient manner.
We therefore hope our work will inspire future research on how best to tap the potential of large foundation models in vision.

\small{
\noindent\textbf{Acknowledgement.}
Menglin is supported by a Meta AI research grant awarded to Cornell University,
Luming and Bharath is supported by NSF IIS-2144117, Serge is supported in part by the Pioneer Centre for AI, DNRF grant number P1.
We would like to thank Alexander Rush, Yin Cui for valuable suggestions and discussion. 
}

\clearpage
\appendix

\section{Implementation Details}
\label{supsec:detail}

We use PyTorch~\cite{paszke2017pytorch} to implement all experiments on NVIDIA A100-40GB GPUs.

\subsection{Classification Experiments}
\subsubsection{VPT.} We use \val{} set of each dataset to find best prompt length $p$, see~\cref{subsec:method_vp}.
The prompt length is the only
VPT-specific hyper-parameter that we tune.
For Transformer backbones, the range of $p$ is $\{1, 5, 10, 50, 100, 200\}$ and $\{1, 5, 10, 50\}$ for \vit{} and \swin{}, respectively. The maximum choice of $p$ is approximately close to the number of image patch tokens 
within each MSA for both architectures (\vit{}: 196, \swin{}: 49).
We also apply a dropout of $0.1$ for~\deepprompt{}.
For ConvNets, the range of $p$ is $\{1, 3, 5, 7, 9, 11\}$.
Each prompt is randomly initialized with xavier uniform initialization scheme~\cite{glorot2010understanding}.
We follow the original backbone' design choices, such as the existence of the classification tokens \texttt{[CLS]}, or whether or not to use the  final \texttt{[CLS]} embeddings for the classification head input.

\subsubsection{\adapter{}.} Adapters~\cite{houlsby2019parameter} insert extra lightweight modules inside each Transformer layer. One adapter module generally consists of a linear down-projection (with a reduction rate $r$), followed by a nonlinear activation function, and a linear up-projection, together with a residual connection. \cite{pfeiffer2020adapterfusion,pfeiffer2020AdapterHub} exhaustively searched all possible configurations and found that only inserting adapters after the FFN ``Add \& LayerNorm'' sub-layer works the best. Therefore we also use this setup in our own implementation. We sweep the reduction rate $r$ in $\{8, 64, 256\}$.

\setlength{\tabcolsep}{4pt}
\begin{table}[!b]
\scriptsize
\begin{center}
\caption{Implementation details for each fine-tuning method evaluated. $\star$: we observe that for \shallowprompt{} sometimes benefit from a larger base LR for 6 out of 24 tasks evaluated, where we search from $\{1000.0, 500.0, 250.0, 100.0\}$
}
\label{table:supp_imp}
\resizebox{\textwidth}{!}{%
\begin{tabular}{l cc}
\toprule
&\fullft{},\partialft{},\bias{},\adapter{} 
&\linear{},\sidetune{}, \mlp{},
\vprompt{}
\\
\midrule
Optimizer &AdamW~\cite{loshchilov2017decoupled} &SGD
\\
Optimizer momentum  &- &0.9 
\\
\emph{base\_lr} range &$\{$0.001, 0.0001, 0.0005, 0.005$\}$ 
&\{50., 25., 10., 5., 2.5, 1.,0.5, 0.25, 0.1, 0.05$\}^{\star}$
\\
Weight decay range &\multicolumn{2}{ c }{$\{0.01, 0.001, 0.0001, 0.0\}$} 
\\
Learning rate schedule &\multicolumn{2}{ c }{cosine decay}
\\
Warm up epochs &\multicolumn{2}{ c }{10}
\\
Total epochs &\multicolumn{2}{ c }{100 (\vit{}-B, \swin{}-B), 50 (\vit{}-L/H)} 
\\
\bottomrule
\end{tabular} %
}
\end{center}
\end{table}
\setlength{\tabcolsep}{1.4pt}

\setlength{\tabcolsep}{4pt}
\begin{table}
\small
\begin{center}
\caption{Specifications of the various datasets evaluated. 
$^{\star}$: we randomly sampled the \texttt{train} and \texttt{val} sets since there are no public splits available
}
\label{table:supp_datasets}
\resizebox{\textwidth}{!}{%
\begin{tabular}{l l  l l l l l}
\toprule
\textbf{Dataset}   &\textbf{Description}  & \textbf{\# Classes}    &\textbf{Train}  &\textbf{Val}  &\textbf{Test} \\ 
\midrule
\multicolumn{3}{l}{Fine-grained visual recognition tasks (FGVC)} 
\\
\cmidrule{2-6}
\quad\cub{}~\cite{WahCUB_200_2011}
& Fine-grained bird species recognition
&200
&5,394$^{\star}$	&600$^{\star}$ &5,794	
\\

\quad\nabirds{}~\cite{van2015nabirds}
& Fine-grained bird species recognition
&55
&21,536$^{\star}$	&2,393$^{\star}$	&24,633
\\

\quad\flowers{}~\cite{nilsback2008automated}
& Fine-grained flower species recognition
&102
&1,020	&1,020	&6,149 
\\

\quad\dogs{}~\cite{Khosla_FGVC2011dogs}
 &Fine-grained dog species recognition  &120 
 &10,800$^{\star}$	&1,200$^{\star}$	&8,580 
\\

\quad\cars{}~\cite{gebru2017cars}
& Fine-grained car classification  &196  
&7,329$^{\star}$	&815$^{\star}$	&8,041 
\\

\midrule

\multicolumn{3}{l}{Visual Task Adaptation Benchmark (\vtab{})~\cite{zhai2019vtab}} 
\\
\cmidrule{2-6}
\quad CIFAR-100~\cite{cifar10} &\multirow{7}{*}{Natural}
&100 &\multirow{7}{*}{800/1000} &\multirow{7}{*}{200} &10,000
\\
  \quad Caltech101~\cite{li2006one} & &102 && &6,084
  \\
  \quad DTD~\cite{cimpoi14describing} & &47 && &1,880
  \\
  \quad Flowers102~\cite{nilsback2008automated} & &102 && &6,149
  \\
  \quad Pets~\cite{parkhi12a} & &37 && &3,669
  \\
  \quad SVHN~\cite{netzer2011reading} & &10 && &26,032
  \\
  \quad Sun397~\cite{xiao2010sun} & &397 && &21,750
  \\
\cmidrule{2-6}

  \quad Patch Camelyon~\cite{veeling2018rotation} &\multirow{4}{*}{Specialized} &2
  &\multirow{4}{*}{800/1000} &\multirow{4}{*}{200} &32,768
  \\
  \quad EuroSAT~\cite{helber2017eurosat} & &10 && &5,400
  \\
  \quad Resisc45~\cite{cheng2017remote} & &45 && &6,300
  \\
  \quad Retinopathy~\cite{kaggle-diabetic-retinopathy} & &5 && &42,670
  \\

\cmidrule{2-6}
  \quad Clevr/count~\cite{johnson2017clevr} &\multirow{8}{*}{Structured}
  &8
  &\multirow{8}{*}{800/1000} &\multirow{8}{*}{200} &15,000
  \\
  \quad Clevr/distance~\cite{johnson2017clevr} & &6 && &15,000
  \\
  \quad DMLab~\cite{beattie2016deepmind} & &6 && &22,735
  \\
  \quad KITTI/distance~\cite{Geiger2013IJRR} & &4 && &711
  \\
  \quad dSprites/location~\cite{dsprites17} & &16 && &73,728
  \\
  \quad dSprites/orientation~\cite{dsprites17} & &16 && &73,728
  \\
  \quad SmallNORB/azimuth~\cite{lecun2004learning} & &18 && &12,150
  \\
  \quad SmallNORB/elevation~\cite{lecun2004learning} & &9 && &12,150
\\

\bottomrule\end{tabular}
}
\end{center}
\end{table}
\setlength{\tabcolsep}{1.4pt}

\setlength{\tabcolsep}{4pt}
\begin{table}
\scriptsize
\begin{center}
\caption{Specifications of different pre-trained backbones used in the paper.
Parameters (M) are of the feature extractor.
``Batch size'' column reports the batch size for \linear{} / \partialft{} / \{\fullft{}, \bias{}, \adapter{}\} / \vprompt{} ($p < 100$) / \vprompt{} ($p \geq 100$). All backbones are pre-trained on \imagenet{}~\cite{imagenet_cvpr09} with resolution 224$\times$224
}
\label{table:supp_backbone}
\resizebox{\textwidth}{!}{%
\begin{tabular}{l l l l l l l}
\toprule
\textbf{Backbone}
&\textbf{\shortstack[l]{Pre-trained\\Objective}}
&\textbf{\shortstack[l]{Pre-trained\\Dataset}}
&\textbf{\shortstack[l]{\# params\\(M)}} 
&\textbf{\shortstack[l]{Feature dim\\$d$}}
&\textbf{\shortstack[l]{Batch Size}}
&\textbf{\shortstack[l]{Pre-trained\\Model}}
\\  
\midrule

\vit-B/16~\cite{dosovitskiy2020vit}

&\multirow{3}{*}{\suplong{}} &\multirow{3}{*}{ImageNet-21k}
& 85 & 768 &2048 / 1280 / 128 / 128 / 64 
&\href{https://storage.googleapis.com/vit_models/imagenet21k/ViT-B_16.npz}{checkpoint}
\\
\vit-L/16~\cite{dosovitskiy2020vit}
& &
&307 & 1024 &  2048 / 640 / 64 / 64 / 32
&\href{https://storage.googleapis.com/vit_models/imagenet21k/ViT-L_16.npz}{checkpoint}
\\
\vit-H/14~\cite{dosovitskiy2020vit}
& &
&630 & 1280 & 1024 / 240 / 28 / 28 / 14
&\href{https://storage.googleapis.com/vit_models/imagenet21k/ViT-H_14.npz}{checkpoint}
\\
\midrule

\vit-B/16~\cite{dosovitskiy2020vit}
&\moco{}~\cite{chen2021mocov3} &\multirow{2}{*}{ImageNet-1k}
&\multirow{2}{*}{85} & \multirow{2}{*}{768} &\multirow{2}{*}{2048 / 1280 / 128 / 128 / 64 }
&\href{https://dl.fbaipublicfiles.com/moco-v3/vit-b-300ep/linear-vit-b-300ep.pth.tar}{checkpoint}
\\
\vit-B/16~\cite{dosovitskiy2020vit}
&\mae{}~\cite{he2021mae}  
&  &  & & 
&\href{https://dl.fbaipublicfiles.com/mae/pretrain/mae_pretrain_vit_base.pth}{checkpoint}
\\
\midrule

\swin-B~\cite{liu2021swin}
&\suplong{}	 &ImageNet-21k
&88	&1024 & 1024 / 1024 / 128 / 80 / -
&\href{https://github.com/SwinTransformer/storage/releases/download/v1.0.0/swin_base_patch4_window7_224_22k.pth}{checkpoint}
\\
\midrule
\rnx-Base~\cite{liu2022convnet} 
&\suplong{} &ImageNet-21k
&88
&1024
&1024 / 1024 / 128 / 128 / -
&\href{https://dl.fbaipublicfiles.com/convnext/convnext_base_22k_224.pth}{checkpoint}
\\
\rn-50~\cite{he2016rn}
&\suplong{} &ImageNet-1k
&23
&2048
&2048 / 2048 / 384 / 256 / -
&\href{https://pytorch.org/vision/stable/models.html}{checkpoint}
\\
\bottomrule
\end{tabular}
} 
\end{center}
\end{table}
\setlength{\tabcolsep}{1.4pt}

\begin{figure}
\centering
\includegraphics[height=0.95\textheight]{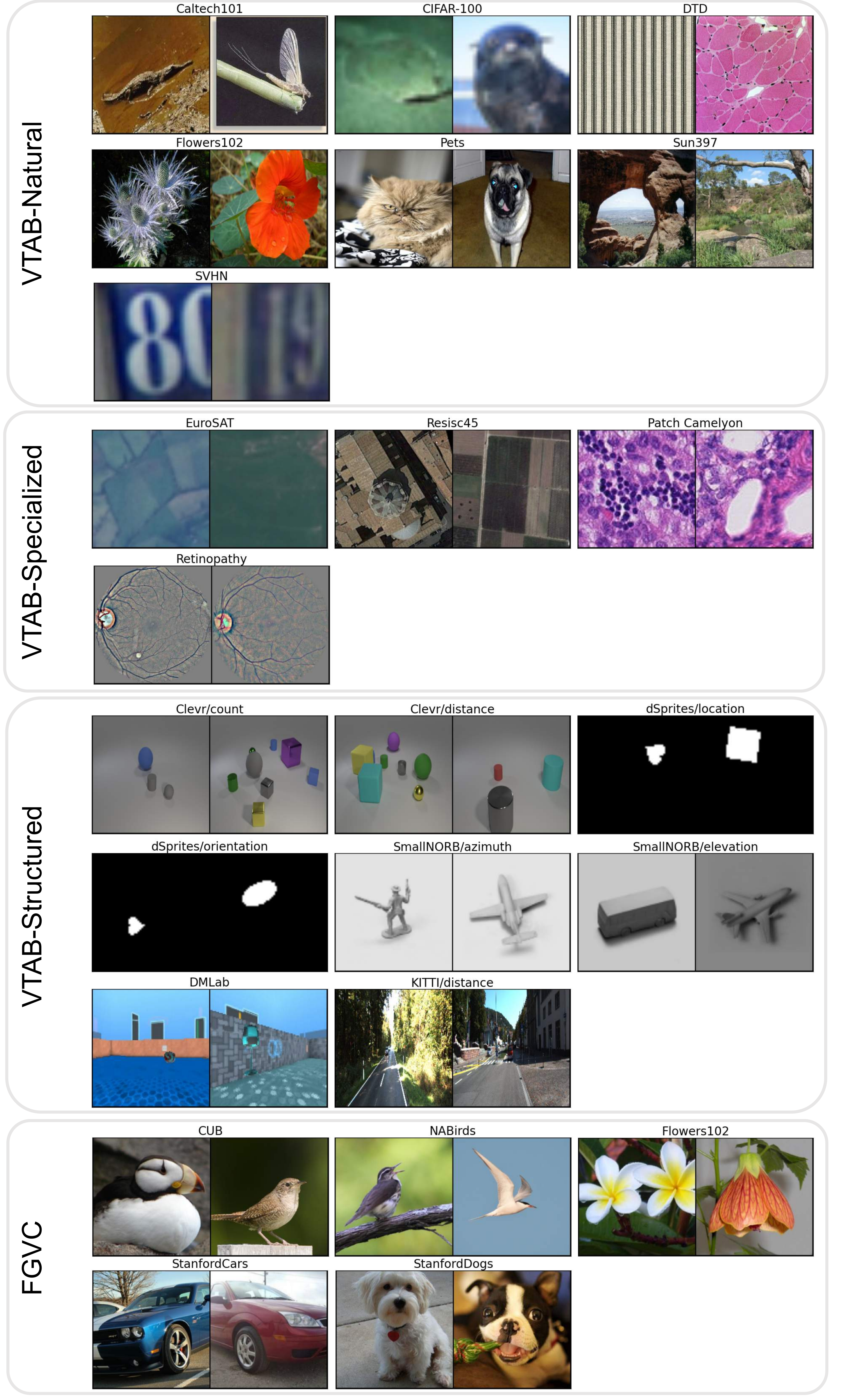}
\caption{Dataset examples for all classification tasks evaluated
}
\label{fig:dataexps}
\end{figure}

\subsubsection{Augmentation and other hyper-parameters.}
We adopt standard image augmentation strategy during training: normalize with \imagenet{} means and standard deviation, randomly resize crop to 224$\times$224 and random horizontal flip for five FGVC datasets, and resize to 224$\times$224 for the \vtab{} suite.\footnote{Following the \href{https://github.com/google-research/task_adaptation/blob/master/task_adaptation/data_loader.py}{default settings} in VTAB, we don't adopt other augmentations}
\cref{table:supp_imp} summarizes the optimization configurations we used.
Following~\cite{wslimageseccv2018}, we conduct grid search to find the tuning-specific hyper-parameters, learning rate, and weight decay values using \val{} set of each task.
Following the linear scaling rule~\cite{krizhevsky2014one,goyal2017accurate,chen2021mocov3,he2021mae},
the learning rate is set as \emph{base\_lr}$\times b / 256$, where $b$ is the batch size used for the particular model, and \emph{base\_lr} is chosen from the range specified in~\cref{table:supp_imp}. 
The optimal hyper-parameter values for each experiment can be found in~\cref{subsec:results_supp}.

\subsubsection{Datasets and pre-trained backbones specifications.}
\cref{table:supp_datasets,table:supp_backbone} summarize the statistics and details of the evaluated classification datasets and all the pre-trained backbones used in the paper.
\cref{fig:dataexps} includes image examples of all 24 classification tasks evaluated.

\subsection{Semantic Segmentation Experiments} 
ADE20K~\cite{zhou2019ade20} is a challenging scene parsing benchmark with 150 fine-grained labels. The training and validation sets contain 20,210 and 2,000 images respectively. We utilize the public codebase MMSegmentation~\cite{mmseg2020} in our implementation.\footnote{See the \href{https://github.com/open-mmlab/mmsegmentation}{MMSegmentation GitHub page}} 
The ViT-L backbone is supervisely pre-trained on \imagenet{}-21k.\footnote{\href{https://storage.googleapis.com/vit_models/augreg/L_16-i21k-300ep-lr_0.001-aug_strong1-wd_0.1-do_0.0-sd_0.0.npz}{ViT-L/16 checkpoint}} 

SETR~\cite{zheng2020rethinking} is a competitive segmentation framework using ViT as the encoder. PUP is a progressive upsampling strategy consisting of consecutive convolution layers and bilinear upsampling operations. Among multiple decoder choices, PUP works the best according to MMSegmentation's reproduction therefore we also use it as in our implementation.\footnote{\href{https://github.com/open-mmlab/mmsegmentation/tree/master/configs/setr}{
MMSegmentation's reproduction on SETR}}

When applying VPT to SETR-PUP, we only insert prompts into SETR's ViT encoder backbone. For the decoder, only image patch embeddings are used as inputs and prompt embeddings are discarded. Same as recognition tasks, only the PUP decoder head and prompts are learned during training and the ViT backbone is frozen.

For full fine-tuning, we use the same hyper-parameters as in MMSegmentation. For \textsc{HeadOnly}, \bias{}, and VPT, we use the hyper-parameter sweep on learning rate \{0.05, 0.005, 0.0005, 0.001\}. The optimal learning rate is 0.005 for all methods.
We sweep prompt length $p\in$ \{1, 5, 10, 50, 100, 200\}. For VPT, we also change the learning rate multiplier to 1.0 instead of the default 10.0, so the decoder head and prompts share the same learning rate.
Other hyper-parameters remain the same as full fine-tuning.

\begin{figure}[t]
\centering
\includegraphics[width=\textwidth]{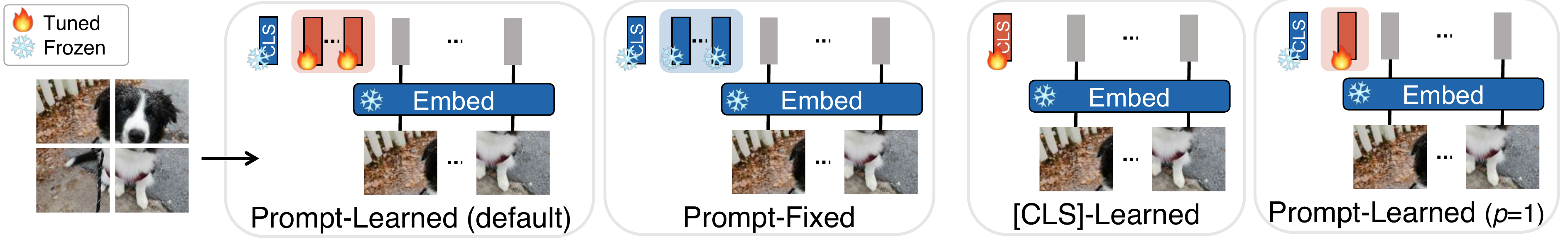}
\includegraphics[width=\textwidth]{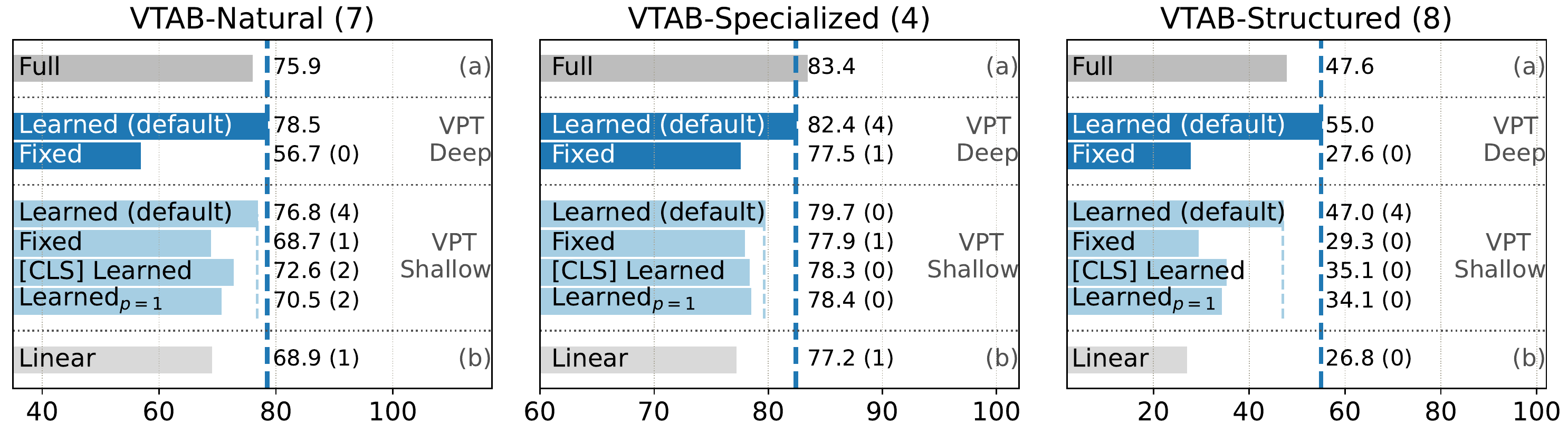}
\caption{Effect of expanding input sequence.
Illustration of different strategies is included at top, and results of those are presented at the bottom section.
For easy comparison, two dark and light blue lines represent the performance of default \deepprompt{} and \shallowprompt{}{}, respectively
}
\label{fig:ablate_update}
\end{figure}

\section{Extended Analysis}
\label{supp_analysis}

\subsubsection{Effect of expanding input sequence length.}
\label{app:seq}
As shown in~\cref{table:main_vitb}, by expanding the input sequence with learnable prompts, VPT achieves better performance than \fullft{} on the 20 out of 24 tasks evaluated.
To investigate whether the advantage of VPT is due to its enlarged input sequence length, 
we experiment on two more variants: (1) the prompts are kept frozen during fine-tuning stage (\texttt{Prompt-Fixed}). (2) only tuning the $\texttt{[CLS]}$ token (\texttt{[CLS]-Learned}).
From \cref{fig:ablate_update} we can see that, updating prompt embeddings (\texttt{Prompt-Learned}) offers
significant gains, while \texttt{Prompt-Fixed} yields comparable results \wrt~\linear{}. 
This suggests that the final performance of VPT is mainly contributed by the learned prompt embeddings instead of the enlarged sequence length. 
Updating the $\texttt{[CLS]}$ token performs similarly as updating 1 prompt ($\texttt{[CLS]}$ \vs{}~\texttt{Learned$_{p=1}$}), but still lags behind the default setting where we manually select the best number of prompt tokens based on the \texttt{val} set.

\begin{figure}[t]
\centering
\includegraphics[width=0.95\textwidth]{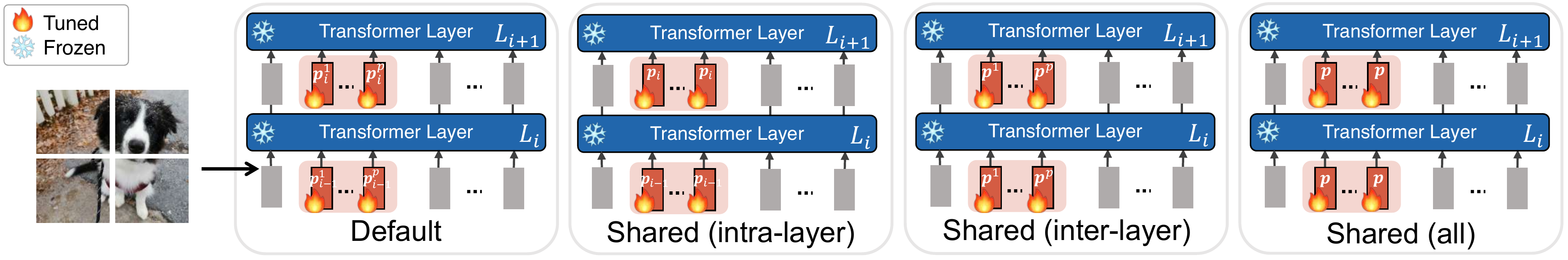}
\includegraphics[width=\textwidth]{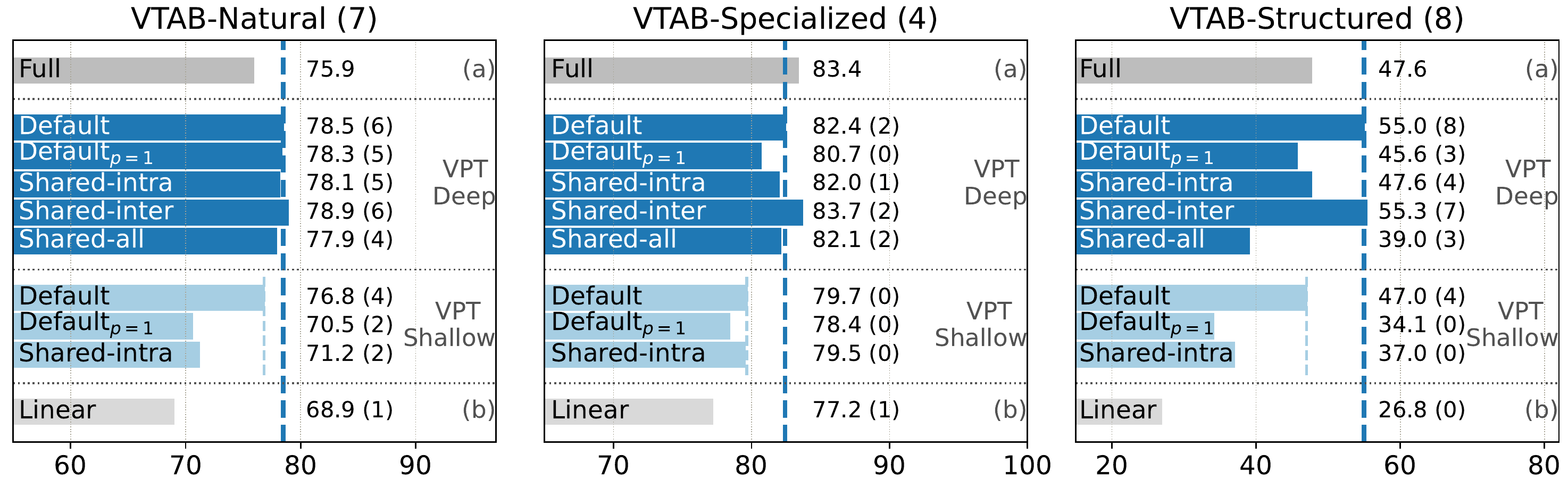}
\caption{Effect of sharing prompts.
Illustration of different strategies is included at top, and results of those are presented at the bottom section.
For easy comparison, the blue dashed line represents the performance of default \deepprompt{}
}
\label{fig:ablate_shared}
\end{figure}

\subsubsection{Sharing prompts.} 
We examine the effect of sharing parameters of prompts in \cref{fig:ablate_shared} by setting the same prompt embedding \emph{within} Transformer layers 
(\texttt{Shared-intra}), \emph{among} all layers 
(\texttt{Shared-inter}), and for all prompts inserted in the Transformer (\texttt{Shared-all}). We can observe that:
(1) Sharing prompts within layer (\texttt{Shared-intra}) performs competitively or slightly outperforms the performance of using one prompt (\texttt{Default$_{p=1}$}), further demonstrating the value of expanding input sequence.
(2) Although \texttt{Shared-intra} under-performs \texttt{Default} in general, surprisingly, \texttt{Shared-inter} slightly outperforms our default \deepprompt{} while using similar number of trainable parameters (total number of parameters for all VTAB tasks: 1.14$\times$ \vs~1.13$\times$ for \texttt{Shared-inter} \vs \texttt{Default}, respectively). Closer examination reveals that the optimal prompt length $p$ for \texttt{Shared-inter} is in general larger than \texttt{Default}, \ie{}, average prompt length on all VTAB tasks: 64.58 \vs~60.94, for \texttt{Shared-inter} \vs \texttt{Default}, respectively.
(3) Sharing the same prompt embedding both among and within layers (\texttt{Shared-all}) deteriorates performance, but still surpass the linear probing results across three VTAB subgroups.

\subsubsection{Prompt initialization.}
In NLP, prompt tuning could 
benefit from more sophisticated prompt initialization, as shown in~\cite{lester-etal-2021-power}. We investigate if this is the case for visual prompting as well.
We utilize prototype representations for downstream target classes so that the prompts are initialized with embeddings that enumerate the output space.
Since we want the model to produce an output embedding that is close to one of these prototype representations given a test example, initializing prompts in this manner might give the model some hints about the target categories thus help improve the optimization process.

Concretely, we use the averaged final \texttt{[CLS]} embeddings whithin each target class of the down-stream dataset \train{} split.
Given the pre-trained ViT with $N$ layers, and the down-stream \train{} set with $c$ target classes, for each training example, we compute the final \texttt{[CLS]} embeddings, $\vec{x}_N \in \R^d$.
Then we average these embeddings within each target class to get $\{\hat{\vec{x}}_N^k\in\R^d \mid k\in\N, 1\le k\le c\}$.\footnote{if $c>200$, we further apply $k$-means ($k=200$) to class-averaged embeddings and use the corresponding 200 centroid embeddings as 
$\{\hat{\vec{x}}_N^k\in\R^d\}_{k=1}^{k=200}$.}
Setting prompt length $p=c$,\footnote{if $c>200$, we set $p=200$ so that prompt length won't be too large. In fact, for VTAB, only the Sun397 task in the \textit{Natural} subgroup has over 200 classes. See~\cref{table:supp_datasets}.}
we initialize $\vec{P}$ with $\{\hat{\vec{x}}_N^k\}_{k=1}^{k=c}$ for \shallowprompt{} , and initialize each $\vec{P}_i$ with $\{\hat{\vec{x}}_N^k\}_{k=1}^{k=c}$, where $i=0, 1, \ldots, N-1$, for \deepprompt{}.

\begin{figure}[t]
\centering
\includegraphics[width=\textwidth]{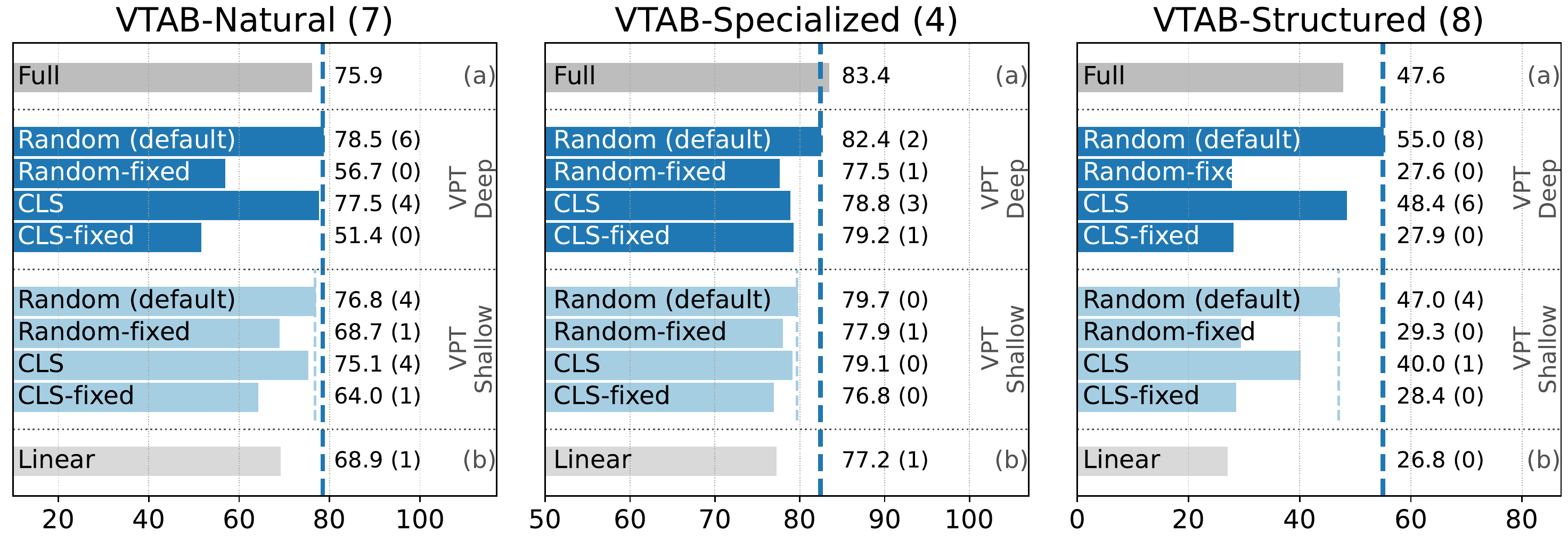}
\caption{Effect of prompt initialization.
For easy comparison, the two blue dashed line represents the performance of default \deepprompt{} and \shallowprompt{}, respectively
}
\label{fig:ablate_init}
\end{figure}

We compare the fine-tuning performance using the above initialization strategy (\texttt{CLS}) against the default random initialization (\texttt{Random}) in~\cref{fig:ablate_init}.
We also report results when we fix the prompts during the fine-tuning stage (\texttt{$\cdot$-fixed}).
As shown in \cref{fig:ablate_init}, it's quite surprising that our default random initialization (\texttt{Random}) works the best in general, consistently across different subgroups of VTAB without extra pre-processing steps described above (\texttt{CLS}).
\texttt{CLS} works comparably in \textit{Natural} and \textit{Specialized} subgroups.\footnote{
Utilizing the per-class averaged \texttt{[CLS]} features, we also tried several other different implementation variants, including using per-layer \texttt{[CLS]} embeddings for \deepprompt{} instead of only the final output \texttt{[CLS]} vector. They perform either the same as or even much worse than the \texttt{CLS} strategy above, and none of them is able to out-perform the default \texttt{Random}.
}

\begin{figure}[t]
\centering
\includegraphics[width=\textwidth]{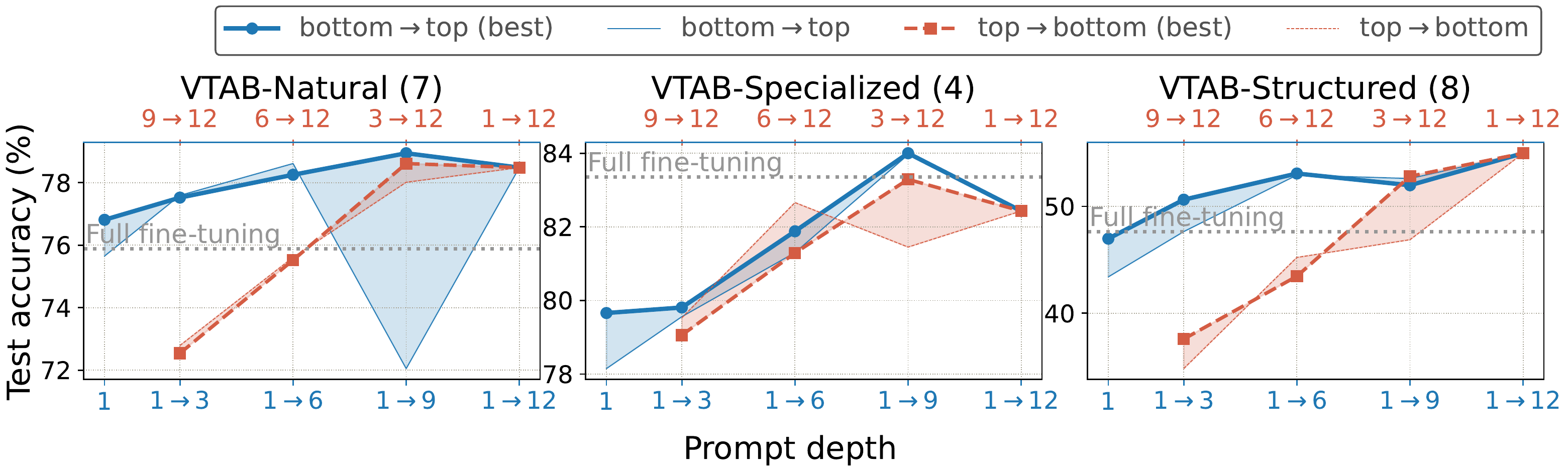}
\caption{Sensitivity to prompt length for the prompt depth experiments.
We select the best prompt length for each variant with \val{} sets.
We also include the same prompt length for all depth choices.
$i\rightarrow j$ indicates the Transformer layer indices that prompts are inserted into. The 1-st layer refers to the one closest to input. ViT-B has a total of 12 layers
}
\label{suppfig:ablate_depth_all}
\end{figure}

\subsubsection{Prompt depth \vs~prompt length.}
In~\cref{fig:ablate_depth}, we ablate the number of layers we insert prompts in. 
For each prompt depth variant, \cref{fig:ablate_depth} reports the results using the best prompt length for \textit{each} 
task (``$\cdot\rightarrow\cdot$ (best)'' in~\cref{suppfig:ablate_depth_all}).
Here we adopt another setting where the best prompt length from $1\rightarrow12$ are used for \textit{all} other prompt depth variants.
Comparing both ``$\cdot\rightarrow\cdot$ (best)'' and ``$\cdot\rightarrow\cdot$'', we observe that there are varied sensitivities to prompt length for different depths, especially if we insert prompts in nine layers only ($3\rightarrow$12, $12\rightarrow3$).

\setlength{\tabcolsep}{4pt}
\begin{table}[t]
\begin{center}
\caption{
Combining \vprompt{} with \bias{} with a pre-trained \vit{}-B in~\cref{subsec:exp_results}.
For each method and each downstream task group,
we report the average test accuracy score and \texttt{number of wins in ($\cdot$)} compared to \fullft{}.
The difference between the hybrid methods and their \vprompt{} counterpart are color coded
}
\label{table:supp_promptbias}
\resizebox{\textwidth}{!}{
\begin{tabular}{
lc!{\color{tabvline}\vrule}
cc !{\color{tabvline}\vrule}
cc}
\toprule
&\bias{}  &\shallowprompt{} &\shallowprompt{} + \bias{}
&\deepprompt{} &\deepprompt{} + \bias{}
\\
\midrule
\bf{VTAB-Natural}
&73.30 (3)  &76.81 (4)  &79.78 (5)\Rise{2.97}  &78.48 (6)  &77.64 (6)\Drop{0.84}
\\
\bf{VTAB-Specialized}
&78.25 (0)  &79.66 (0)  &81.38 (0)\Rise{1.72}  &82.43 (2)  &82.22 (2)\Drop{0.21}
\\
\bf{VTAB-Structured}
&44.09 (2)  &46.98 (4)  &45.89 (3)\Drop{1.09}  &54.98 (8)  &53.87 (6)\Drop{1.11}
\\
\bottomrule
\end{tabular}
}
\end{center}
\end{table}
\setlength{\tabcolsep}{1.4pt}

\begin{figure}[t]
\centering
\includegraphics[width=\textwidth]{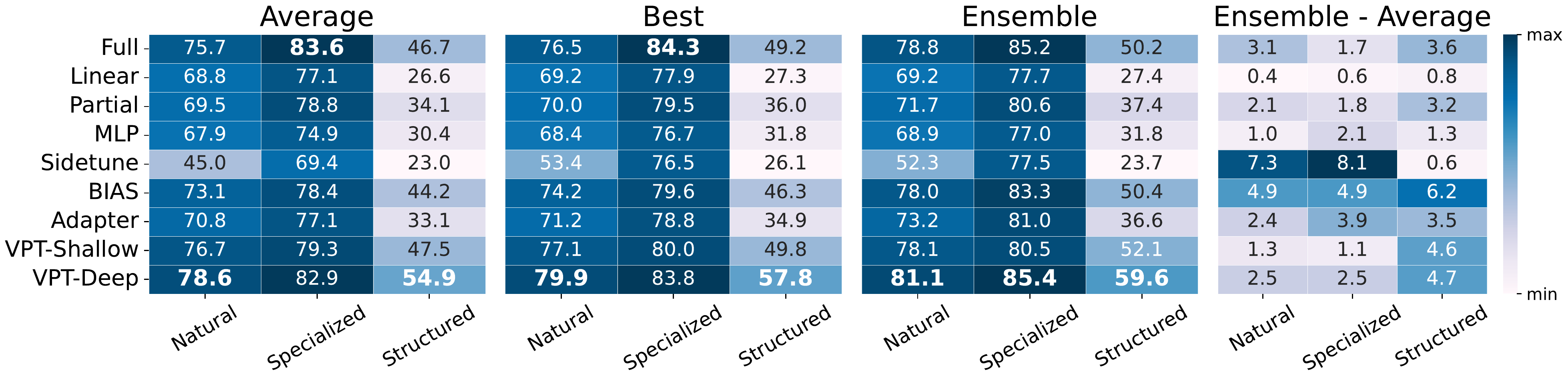}
\caption{
Performance of a five-run ensemble. We report the averaged, the best among five runs as well.
Best performance is bolded in each column
}
\label{fig:ensemble}
\end{figure}

\subsubsection{Combine VPT with Bias Tuning.}
Our experiments in the main paper reveal that \bias{} is a competitive parameter-efficient tuning baseline (\eg,~\cref{table:main_vitb}\texttt{(c)}). 
Based on this observation, we explore another protocol where we update both prompts and the bias terms of the pre-trained backbone, keeping everything else in the backbone frozen (\promptbias{}). 
As shown in \cref{table:supp_promptbias}, to our surprise, incorporating \bias{} with \vprompt{} does not yield superior results in general, even undermines \deepprompt{} for all 3 task subgroups. This suggests that these two methods are not necessarily complementary to each other.

\subsubsection{Prompt ensembling.}
\cite{lester-etal-2021-power} demonstrated prompt's efficiency in the context of model ensembling. 
For an ensemble of $k$ models, we only need to store the learnt prompt vectors instead of $k$ copies of the whole fine-tuned model parameters (\eg{}, $k\times2.5$GB for ViT-H).
Furthermore, given one test example during inference time, \emph{only one} forward pass is executed with a specially-designed batch with replicated original data but varied prompts.

Given such advantages, we also investigate VPT's effectiveness on prompt ensembling. We train 5 different prompts for each VTAB task with different random seeds, using the same pre-trained ViT-B backbone and hyper-parameters as in \cref{table:main_vitb}. \cref{fig:ensemble} shows that the ensembled \deepprompt{} outperforms the average or even the best single-prompt counterparts, as well as other ensembled fine-tuning methods including \fullft{}.

\setlength{\tabcolsep}{4pt}
\begin{table}[t]
\begin{center}
\caption{Non-parametric paired one-tailed $t$-test  (the Wilcoxon signed-rank test) on whether \deepprompt{}'s performance is greater than other methods on 19 VTAB tasks. Results show that,
\deepprompt{} is indeed statistically significantly better than other fine-tuning protocols ($p<0.05$)
}
\label{table:supp_sig}
\resizebox{\textwidth}{!}{
\begin{tabular}{
l c !{\color{tabvline}\vrule}
ccc   !{\color{tabvline}\vrule}
cc c !{\color{tabvline}\vrule}
c}
\toprule
&\ttbf{(a)} 
&\multicolumn{3}{c!{\color{tabvline}\vrule}}{\ttbf{(b)} }
&\multicolumn{3}{c!{\color{tabvline}\vrule}}{\ttbf{(c)}}
&\ttbf{(ours)} 
\\
&\fullft{} &\linear{} &\mlp{}-3 &\partialft{}-1 &\sidetune{} &\bias{} &\adapter{} &\shallowprompt{}
\\
\midrule
\shortstack[l]{Is \deepprompt{} statistically \\significantly better?}
&\checkmark
&\checkmark &\checkmark &\checkmark 
&\checkmark &\checkmark &\checkmark 
&\checkmark
\\
$p$-value 

& 1.2e-03
& 2.7e-05
& 1.9e-06
& 1.9e-05
& 1.9e-06
& 1.9e-06
& 3.8e-06
& 2.7e-05
\\
\bottomrule

\end{tabular}
}
\end{center}
\end{table}
\setlength{\tabcolsep}{1.4pt}

\begin{figure}[t]
\centering
\includegraphics[width=\textwidth]{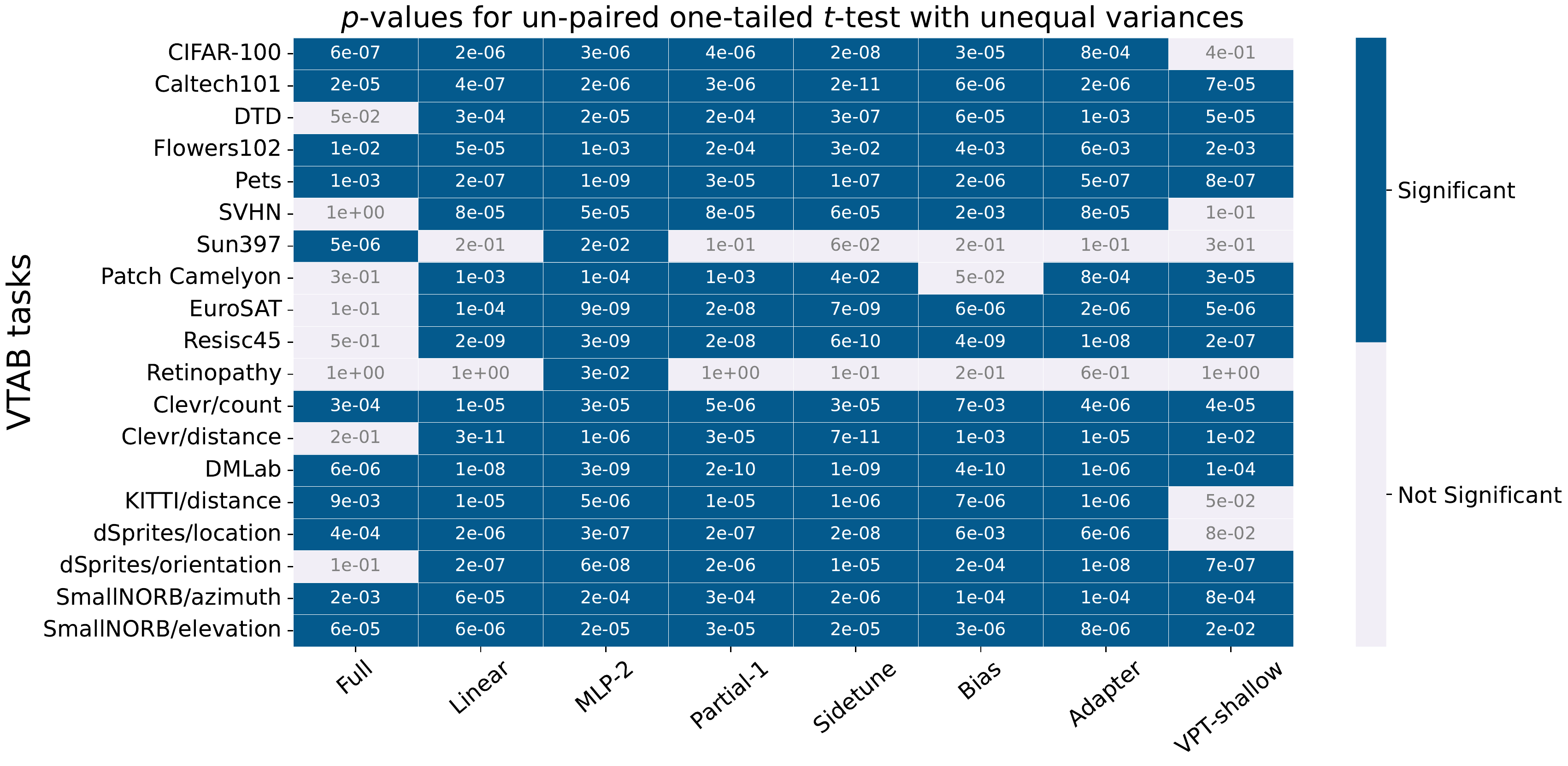}
\caption{Un-paired one-tailed $t$-test with unequal variances (Welch's $t$-test) on whether \deepprompt{}'s performance is greater than other methods for each VTAB task.
Results show that,
\deepprompt{} is statistically significantly better than other fine-tuning protocols ($p<0.05$) in most instances
}
\label{suppfig:sig}
\end{figure}

\begin{figure}
\centering
\includegraphics[width=\textwidth]{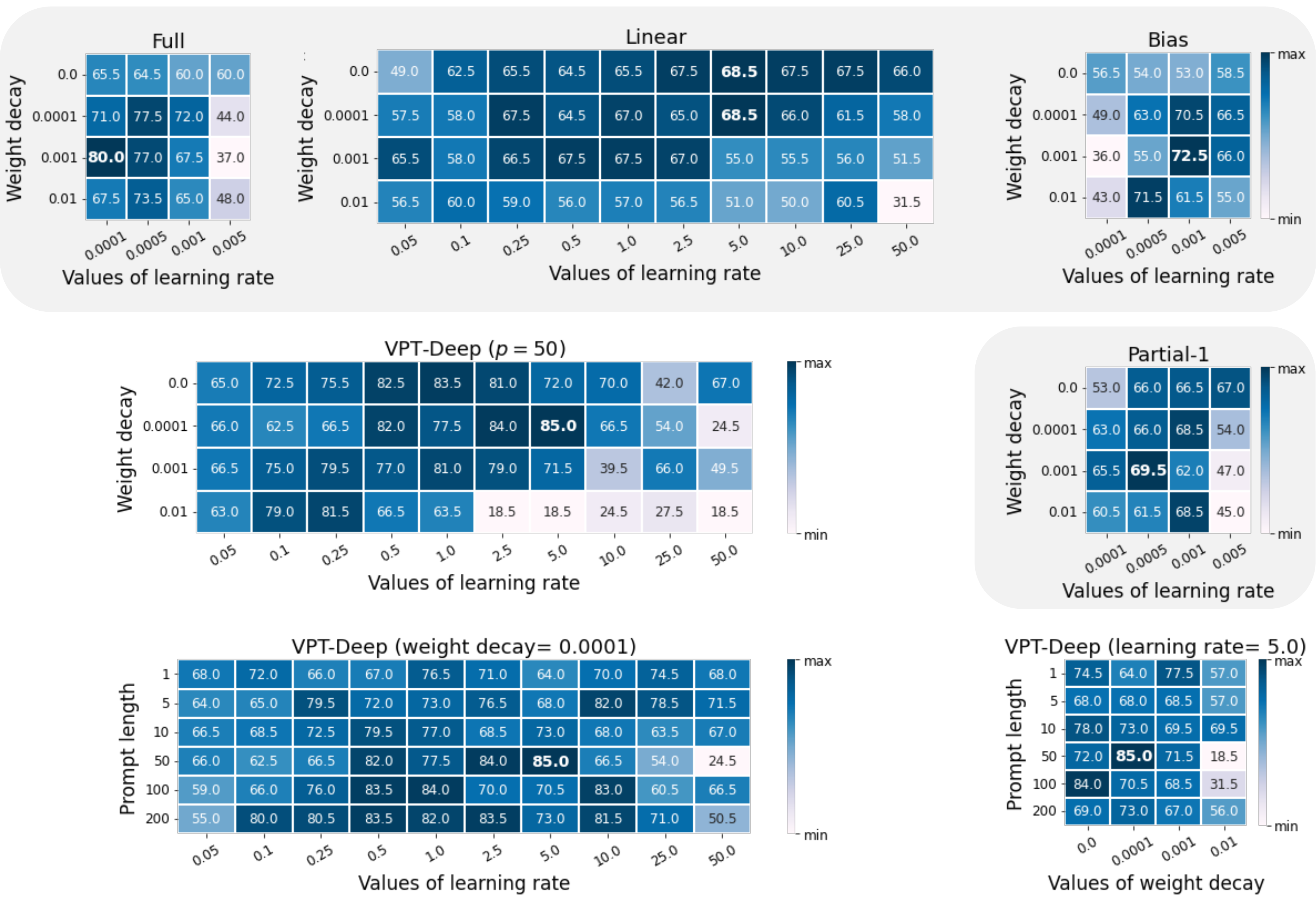}
\caption{Effect of different fine-tuning hyperparameters. Evaluated on the VTAB-\textit{Specialized}: KITTI/Distance task. Other tuning methods are shaded in gray
}
\label{fig:hp}
\end{figure}

\subsubsection{Test of statistical significance.} We conduct non-parametric paired one-tailed $t$-test (the Wilcoxon signed-rank test~\cite{wilcoxon1992individual}) on whether \deepprompt{}'s performance is greater than other fine-tuning methods across 19 VTAB tasks (the null hypothesis $H_0$ states that the mean VTAB performance difference between \deepprompt{} and alternate baseline method is zero.
The alternative hypothesis $H_1$ states that \deepprompt{} outperforms the baseline method on VTAB).
\cref{table:supp_sig} presents the $p$-values of each test, with the number of observations equal to 19 for each method compared (we use the averaged accuracy scores among 5 runs for 19 VTAB tasks and all fine-tuning methods).
For all of the fine-tuning protocols compared, 
\deepprompt{}'s improvements are statistically significant ($p<0.05$).

We also conduct un-paired one-tailed $t$-test with unequal variances (Welch's $t$-test~\cite{welch1947generalization}), comparing the individual runs (the number of observations = 5) for each VTAB task 
($H_0$ states that \deepprompt{} and the other baseline perform the same for a specific VTAB task, while $H_1$ states that \deepprompt{} outperforms the other baseline for a specific VTAB task).
\cref{suppfig:sig} presents the $p$-values for each $<$\deepprompt{}, baseline method$>$ pair on each task.
We reject $H_0$ on 127 out of $19\times 8=152$ cases ($p<0.05$).
Compared to \fullft{}, \deepprompt{} achieves statistically significant better performance on 11 out of 19 tasks.

\setlength{\tabcolsep}{4pt}
\begin{table}[htpb]
\begin{center}
\caption{\vit{}-B/16 pre-trained on supervised \imagenet-21k, fine-tuned with resolution 384$\times$384. We also include \vprompt{} with image resolution 224$\times$224, $p=380$, so the effective image resolution is 384$\times$384.
For each method and each downstream task group,
we report the average test accuracy score and \texttt{number of wins in ($\cdot$)} compared to \fullft{}.
``Total params'' denotes total parameters needed for all 24 downstream tasks. 
Best results among all methods except \fullft{} are \textbf{bolded}
}
\label{table:supp_384}
\resizebox{\textwidth}{!}{
\begin{tabular}{ll r r rrr}
\toprule
&\textbf{\vit{}-B/16 } &\bf{Fine-tune} &\bf{Total}
&\multicolumn{3}{c}{\bf{\vtab{}}}
\\
&\bf{(85.8M)} &\bf{Resolution}  
&\bf{params}
&\bf{Natural}
&\bf{Specialized}
&\bf{Structured}
\\
\midrule
&Total \# of tasks& &
 &7 &4 &8
\\
\midrule
\band \multirow{2}{*}{\ttbf{(a)}}&\fullft{} &384 & 19.07$\times$  
&72.57  &83.05 &50.86
\\
&\fullft{} &224 & 19.07$\times$  
&75.88 &83.36 &47.64
\\
\midrule
\multirow{3}{*}{\ttbf{(b)}}&\linear{} & &1.01$\times$ 
&66.30 (2) &76.77 (0) &27.86 (0)
\\
&\mlp{}-3 &384 &1.27$\times$  
&66.45 (3) &77.77 (0) &38.03 (0)
\\
&\partialft{}-1  & &2.58$\times$  
&67.91 (4) &76.94 (0) &37.16 (0)
\\
\midrule
&\sidetune{} &&3.12$\times$  
&47.08 (1) &40.34 (0) &24.18 (0)
\\
\ttbf{(c)}&\bias{} &384 &1.03$\times$ 
&70.30 (4) &76.06 (0) &45.35 (1)
\\
&\adapter{} & &1.11$\times$ 
&69.42 (6) &77.11 (0) &30.62 (0)
\\
\midrule
\multirow{6}{*}{\ttbf{(ours)}}
&\shallowprompt{} ($p\in\{1, 5, 10, 50, 100, 200\}$) &384 & 1.02$\times$ 
 &75.30 (4) &78.50 (0) &46.56 (2)
\\
&\deepprompt{} ($p\in\{1, 5, 10, 50, 100, 200\}$) 
&384 &1.19$\times$ 
&\textbf{79.37} (\textbf{6}) &\textbf{82.86} (\textbf{2}) &\textbf{56.36} (\textbf{7})
\\
\cmidrule{2-7}
&\shallowprompt{} ($p=380$) &224 & 1.07$\times$ 
 &75.07 (3) &79.03 (0) &46.21 (2)
\\
&\deepprompt{} ($p=380$) &224 &1.78$\times$ 
&74.20 (4) &82.30 (2) &54.50 (6)
\\
\bottomrule
\end{tabular}
}
\end{center}
\end{table}
\setlength{\tabcolsep}{1.4pt}

\subsubsection{Effect of different fine-tuning hyper-parameters.}
In \cref{fig:hp}, we present different tuning protocol's performance on different fine-tuning hyper-parameters including learning rate and weight decay. For our proposed \deepprompt{}, we also ablate different choices of prompt length $p$, which is the only hyper-parameter that needs to be manually tuned.
All experiments are 
evaluated on the \val{} set of KITTI/Distance task (VTAB-\textit{Specialized}). 
We observe different behaviors between \linear{} and \vprompt{}. Both methods freeze backbone parameters during fine-tuning stage.
Linear probing is more sensitive to weight decay values in general, whereas VPT is influenced by both learning rate and weight decay values.
VPT with larger prompt length is also less sensitive to the choice of learning rate.

\subsubsection{Effect of image resolution.}
The original ViT paper~\cite{dosovitskiy2020vit} found that fine-tuning with higher image resolutions (384$\times$384) is beneficial to downstream recognition tasks. All recognition experiments presented in the main paper are fine-tuned on 224$\times$224 resolution. 
As shown in \cref{table:supp_384}, we re-run the VTAB experiments with the same setup as in \cref{table:main_vitb} but in the 384 resolution instead of the default 224. We can see that, \deepprompt{} still achieves the best performance among all parameter-efficient tuning protocols, and even outperforms full fine-tuning on 15 out of 19 tasks. 
Although the increase of image resolutions doesn't lead to better full fine-tuning performance in general, it indeed slightly boosts \deepprompt{}'s performance.

Another interesting observation from~\cref{table:supp_384} is that with 224 fine-tune resolution and a larger value of 
$p=380$, \vprompt{} could achieve similar or better performance compared to \fullft{} with 384 resolution, while using the same input sequence length yet significantly less trainable parameters.

\subsubsection{Empirical computational cost.}
\begin{figure}[t]
\centering
\includegraphics[width=0.9\textwidth]{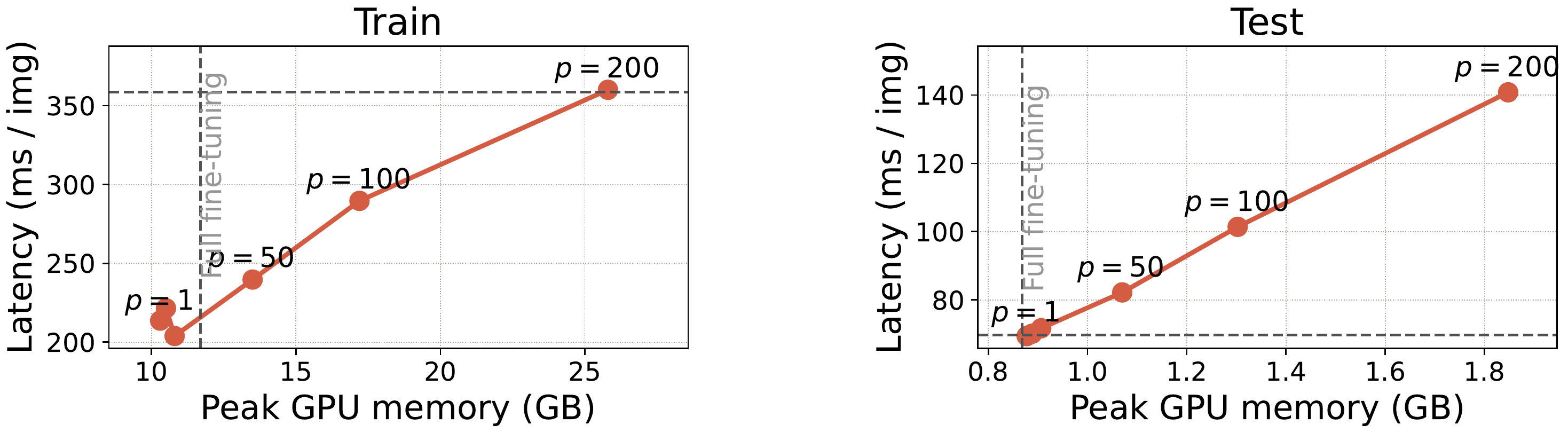}
\caption{Peak GPU memory and latency (ms/img) during both training (left) and inference time (right).
For easy comparison, the gray dashed lines represent latency and memory of full fine-tuning
}
\label{suppfig:cost}
\end{figure}

One possible limitation of \vprompt{} is the extra input sequence length for Transformers. In theory the complexity of MSA is quadratic \wrt{} the input sequence length, but this might not be the case for real-world speed due to hardware details like lane widths and cache sizes~\cite{dosovitskiy2020vit}.
In~\cref{table:supp_cost,suppfig:cost}, we study the empirical computational cost, 
\ie{}, latency, and peak GPU memory usage at both training and inference times, for all the fine-tuning protocols studied. All experiments use the same A100 GPU with a batch size 64 for both training and inference.
We can see that the theoretical quadratic scaling \wrt{} sequence length barely happens to VPT. For instance, doubling the length ($p=200$~\vs{}~$m=198$) basically only lead to 2$\times$ (instead of 4$\times$) inference latency and peak GPU memory \wrt{} full fine-tuning. For training, the latency would be largely reduced with less number of prompts.

\begin{figure}
\centering
\includegraphics[width=0.9\textwidth]{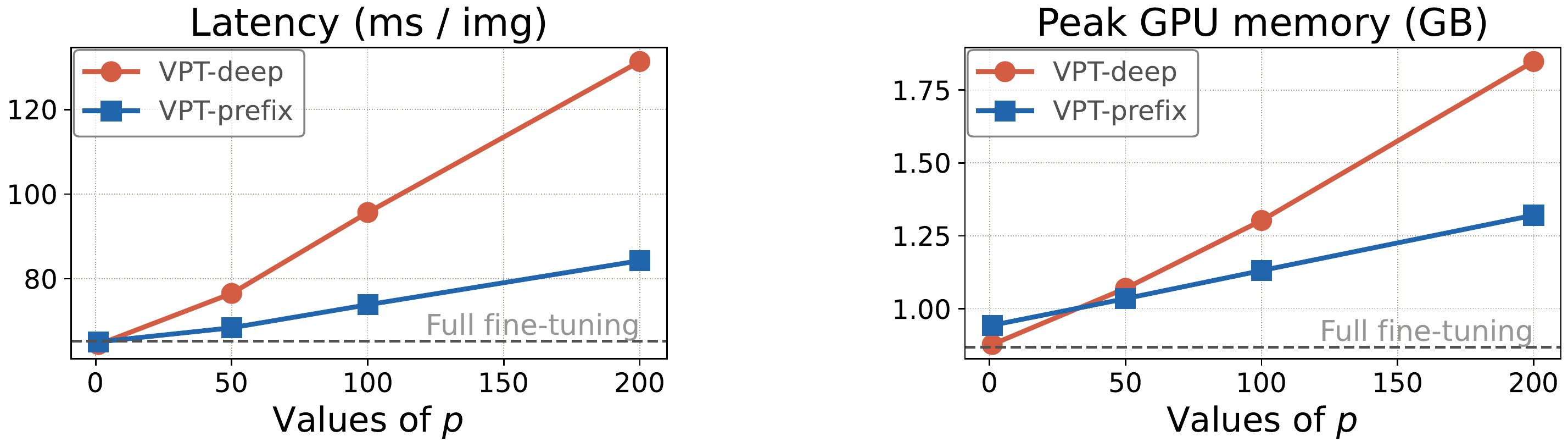}
\caption{VPT-deep \vs~VPT-prefix: peak GPU memory (left) and latency (right) during inference time.
For easy comparison, the gray dashed lines represent latency and memory of full fine-tuning
}
\label{suppfig:cost_vpt_prefix}
\end{figure}

An equivalent implementation of \vprompt{} during test time is directly prepend the parameters to the key and value arrays inside the self-attention module of Transformer~\cite{li-liang-2021-prefix} (\vprompt{}-prefix). While we found that such implementation does not lead to accuracy improvement on VTAB datasets, it reduces the computation cost during inference.
\Cref{suppfig:cost_vpt_prefix} shows the comparison with different values of $p$. \vprompt{}-prefix reduces test-time latency and peak GPU memory 
with a large margin especially when $p$ becomes large.

\setlength{\tabcolsep}{4pt}
\begin{table}[t]
\caption{Cost analysis using a \vit{}-B/16 pre-trained on supervised \imagenet-21k. 
For each method and each downstream task group,
we report the latency (\textbf{ms/img}) and peak GPU memory usage (\textbf{GB}) at both training and inference time.
``Tuned params'' denotes the fraction of learnable parameters needed. 
``Scope'' denotes the tuning scope of each method.
``Extra params'' denotes the presence of additional parameters
besides the pre-trained backbone and linear head.
All experiments use the same A100 GPU
}
\begin{center}
\label{table:supp_cost}
\resizebox{\textwidth}{!}{
\begin{tabular}{
ll  !{\color{tabvline}\vrule}
r   !{\color{tabvline}\vrule}
cc c !{\color{tabvline}\vrule}
c !{\color{tabvline}\vrule}
rr !{\color{tabvline}\vrule}
rr }
\toprule
&\textbf{\vit{}-B/16 }
&\bf{Tuned}
&\multicolumn{3}{c!{\color{tabvline}\vrule}}{\bf{Scope}}
&\bf{Extra}
&\multicolumn{2}{c!{\color{tabvline}\vrule}}{\bf{Train}}
&\multicolumn{2}{c}{\bf{Test}}

\\
&\bf{(85.8M)}
&\bf{params}
&\bf{Input} &\bf{Backbone}&\bf{Head} &\bf{params}
&\bf{Latency}  &\bf{Memory} &\bf{Latency}  &\bf{Memory}
\\
&&&&&&&\bf{\scriptsize{(ms/img)}} &\bf{\scriptsize{(GB)}} &\bf{\scriptsize{(ms/img)}} &\bf{\scriptsize{(GB)}}
\\
\midrule
\band \ttbf{(a)}&\fullft{} & 100$\%$ & &\checkmark  &\checkmark &
&358.7 &11.7 &69.7 &0.87
\\
\midrule
\multirow{3}{*}{\ttbf{(b)}}
&\linear{} &0.09$\%$ & &  & & 
&148.9 &0.9 &64.4 &0.87
\\
&\partialft{}-1 &8.35$\%$ & &  &\checkmark &
&193.2 &1.4 &66.1 &0.87
\\
 &\mlp-3 &1.45$\%$ &&  & &\checkmark
&164.3 &0.9 &64.4 &0.87
\\

\midrule

\multirow{5}{*}{\ttbf{(c)}}&\sidetune & 10.09$\%$ &&\multirow{5}{*}{\checkmark} & &\checkmark
&164.6 &1.2 &66.9 &0.91
\\
&\bias{} &0.21$\%$ && & &
&296.9 &10.1 &65.6 &0.87
\\
&\adapter{} ($r=8$) &2.12$\%$ && & &\checkmark
&293.4 &9.9 &68.2 &0.87
\\
&\adapter{} ($r=64$) &0.36$\%$ && & &\checkmark
&294.4 &9.8 &68.3 &0.87
\\
&\adapter{} ($r=256$) &0.17$\%$ && & &\checkmark
&271.4 &9.8 &68.0 &0.87
\\
\midrule
\multirow{4}{*}{\ttbf{(ours)}}
&\shallowprompt{} ($p=1$) & 0.09$\%$ &\multirow{4}{*}{\checkmark} & & &\multirow{4}{*}{\checkmark}
&205.9 &10.3 &68.1 &0.88
\\
&\deepprompt{} ($p=1$) &0.10$\%$ & && &
&213.6 &10.3 &69.4 &0.88
\\
&\shallowprompt{} ($p=200$) & 0.27$\%$ & & & &
&350.6 &25.8 &138.8 &1.84
\\
&\deepprompt{} ($p=200$) &2.19$\%$ & && &
&360.1 &25.8 &140.8 &1.85
\\
\bottomrule

\end{tabular}
}
\end{center}
\end{table}
\setlength{\tabcolsep}{1.4pt}

\section{Further Discussion}
\subsubsection{VPT \vs~Adversarial Reprogramming (AR).}
\label{vpt_ar}
The differences are:
(1) the number of learnt parameters injected in the input space in AR literature~\cite{elsayed2018adversarial} is nearly 20 times larger than ours (264k \vs 13k). VPT is significantly more parameter-efficient;
(2) AR has shown its effectiveness in ConvNet, while VPT can be applied to broader architectures, including ViT, Swin. Furthermore, VPT is more general with the option of diving into deeper layers of pre-trained backbone (\cref{fig:method}), whereas AR strictly applies to the \emph{first} input layer of ConvNets.
(3) another distinction is that our setting update both prompts and classification head, while AR~\cite{elsayed2018adversarial} directly use the pre-trained classification head. Our setup is more general and could be applied to models with a broader range of pre-training objectives (\eg, MAE~\cite{he2021mae}, which does not include a pre-trained classification head) and  broader vision tasks (\eg, segmentation).

\subsubsection{Visual prompt \vs~textual prompt.}
Our paper also discover discrepancies between visual and textual prompts: we show that VPT could even outperform full-model fine-tuning on 20 out of 24 cases, which is in contract to the NLP's related work~\cite{lester-etal-2021-power}. We also found that random initialized prompts works better in~\cref{fig:ablate_init},
and prompts at earlier layers matters more (\cref{fig:ablate_depth,suppfig:ablate_depth_all}), which are also different from observation on the NLP side~\cite{lester-etal-2021-power,liu2021p}.
These discrepancies indicate that visual prompting might be fundamentally different from text prompts thus in need of further investigation.

\section{Supplementary Results}
\label{subsec:results_supp}

\setlength{\tabcolsep}{4pt}
\begin{table}[b]
\scriptsize
\caption{Per-task fine-tuning results from~\cref{table:main_vitb} for \vtab{} with a pre-trained \vit{}-B/16
}
\label{table:supp_vtab}
\resizebox{\textwidth}{!}{
\begin{tabular}{
c l 
rrrrrrr r!{\color{tabvline}\vrule}
rrrr r!{\color{tabvline}\vrule}
rrrrrrrr r
}
\toprule
  &&\rotatebox{90}{\bf{CIFAR-100}}
  &\rotatebox{90}{\bf{Caltech101} }
  &\rotatebox{90}{\bf{DTD} }
  &\rotatebox{90}{\bf{Flowers102} }
  &\rotatebox{90}{\bf{Pets} }
  &\rotatebox{90}{\bf{SVHN} }
  &\rotatebox{90}{\bf{Sun397} }
  &\rotatebox{90}{\bf{Mean}}
  &\rotatebox{90}{\bf{Patch Camelyon} }
  &\rotatebox{90}{\bf{EuroSAT} }
  &\rotatebox{90}{\bf{Resisc45} }
  &\rotatebox{90}{\bf{Retinopathy} }
  &\rotatebox{90}{\bf{Mean}}
  &\rotatebox{90}{\bf{Clevr/count} }
  &\rotatebox{90}{\bf{Clevr/distance} }
  &\rotatebox{90}{\bf{DMLab}}
  &\rotatebox{90}{\bf{KITTI/distance} }
  &\rotatebox{90}{\bf{dSprites/location} }
  &\rotatebox{90}{\bf{dSprites/orientation} }
  &\rotatebox{90}{\bf{SmallNORB/azimuth} }
  &\rotatebox{90}{\bf{SmallNORB/elevation} }
  &\rotatebox{90}{\bf{Mean}}
  \\
\midrule
\band \ttbf{(a)}&\fullft{} &68.9 &87.7 &64.3 &97.2 &86.9 &87.4 &38.8 &75.88 &79.7 &95.7 &84.2 &73.9 &83.36 &56.3 &58.6 &41.7 &65.5 &57.5 &46.7 &25.7 &29.1 &47.64 
\\
\midrule
\multicolumn{4}{ l }{\emph{Head-oriented}} 
\\
\multirow{6}{*}{\ttbf{(a)}}
&\linear{} &63.4 &85.0 &63.2 &97.0 &86.3 &36.6 &51.0 &68.93 (1) &78.5 &87.5 &68.6 &74.0 &77.16 (1) &34.3 &30.6 &33.2 &55.4 &12.5 &20.0 &9.6 &19.2 &26.84 (0)
\\
&\partialft{}-1 &66.8 &85.9 &62.5 &97.3 &85.5 &37.6 &50.6 &69.44 (2) &78.6 &89.8 &72.5 &73.3 &78.53 (0) &41.5 &34.3 &33.9 &61.0 &31.3 &32.8 &16.3 &22.4 &34.17 (0)
\\
&\mlp{}-2 &63.2 &84.8 &60.5 &97.6 &85.9 &34.1 &47.8 &67.70 (2) &74.3 &88.8 &67.1 &73.2 &75.86 (0) &45.2 &31.6 &31.8 &55.7 &30.9 &24.6 &16.6 &23.3 &32.47 (0)
\\
&\mlp{}-3 &63.8 &84.7 &62.3 &97.4 &84.7 &32.5 &49.2 &67.80 (2) &77.0 &88.0 &70.2 &56.1 &72.83 (0) &47.8 &32.8 &32.3 &58.1 &12.9 &21.2 &15.2 &24.8 &30.62 (0)
\\
&\mlp{}-5 &59.3 &84.4 &59.9 &96.1 &84.4 &30.9 &46.8 &65.98 (1) &73.7 &87.2 &64.8 &71.5 &74.31 (0) &50.8 &32.3 &31.5 &56.4 &7.5 &20.8 &14.4 &20.4 &29.23 (0)
\\
&\mlp{}-9 &53.1 &80.5 &53.9 &95.1 &82.6 &24.4 &43.7 &61.90 (1) &78.5 &83.0 &60.2 &72.3 &73.49 (0) &47.5 &27.9 &28.9 &54.0 &6.2 &17.7 &10.8 &16.2 &26.15 (0)
\\
\midrule
\multicolumn{4}{ l }{\emph{Backbone-oriented}} 
\\
\multirow{6}{*}{\ttbf{(b)}}
&\sidetune{} &60.7 &60.8 &53.6 &95.5 &66.7 &34.9 &35.3 &58.21 (0) &58.5 &87.7 &65.2 &61.0 &68.12 (0) &27.6 &22.6 &31.3 &51.7 &8.2 &14.4 &9.8 &21.8 &23.41 (0)
\\
&\bias{} &72.8 &87.0 &59.2 &97.5 &85.3 &59.9 &51.4 &73.30 (3) &78.7 &91.6 &72.9 &69.8 &78.25 (0) &61.5 &55.6 &32.4 &55.9 &66.6 &40.0 &15.7 &25.1 &44.09 (2)
\\
&\adapter{}-256 &74.1 &86.1 &63.2 &97.7 &87.0 &34.6 &50.8 &70.50 (4) &76.3 &88.0 &73.1 &70.5 &76.98 (0) &45.7 &37.4 &31.2 &53.2 &30.3 &25.4 &13.8 &22.1 &32.39 (0)
\\
&\adapter{}-64 &74.2 &85.8 &62.7 &97.6 &87.2 &36.3 &50.9 &70.65 (4) &76.3 &87.5 &73.7 &70.9 &77.10 (0) &42.9 &39.9 &30.4 &54.5 &31.9 &25.6 &13.5 &21.4 &32.51 (0)
\\
&\adapter{}-8 &74.2 &85.7 &62.7 &97.8 &87.2 &36.4 &50.7 &70.67 (4) &76.9 &89.2 &73.5 &71.6 &77.80 (0) &45.2 &41.8 &31.1 &56.4 &30.4 &24.6 &13.2 &22.0 &33.09 (0)
\\
\midrule
\multicolumn{4}{ l }{\emph{Visual-Prompt Tuning}} 
\\
\multirow{6}{*}{\ttbf{(ours)}}&\shallowprompt{} &77.7 &86.9 &62.6 &97.5 &87.3 &74.5 &51.2 &76.81 (4) &78.2 &92.0 &75.6 &72.9 &79.66 (0) &50.5 &58.6 &40.5 &67.1 &68.7 &36.1 &20.2 &34.1 &46.98 (4)
 \\
 &Prompt length ($p$) &100 &5 &1 &200 &50 &200 &1 &79.4 &5 &50 &50 &10 &28.7 &100 &200 &100 &100 &100 &100 &200 &200 &137.5
 \\
&Tuned / Total (\%) &0.18 &0.10 &0.04 &0.27 &0.08 &0.19 &0.36 &0.17 &0.01 &0.05 &0.09 &0.01 &0.04 &0.10 &0.18 &0.09 &0.09 &0.10 &0.10 &0.19 &0.19 &0.13
\\
\cmidrule{2-24}
&\deepprompt{} &78.8 &90.8 &65.8 &98.0 &88.3 &78.1 &49.6 &78.48 (6) &81.8 &96.1 &83.4 &68.4 &82.43 (2) &68.5 &60.0 &46.5 &72.8 &73.6 &47.9 &32.9 &37.8 &54.98 (8)
\\
&Prompt length ($p$) &10 &10 &10 &1 &1 &50 &5 &12.4 &100 &100 &10 &1 &52.8 &50 &200 &100 &50 &10 &50 &200 &200 &107.5
\\
&Tuned / Total (\%) &0.20 &0.20 &0.15 &0.10 &0.04 &0.54 &0.41 &0.23 &1.06 &1.07 &0.15 &0.02 &0.57 &0.54 &2.11 &1.07 &0.54 &0.12 &0.55 &2.12 &2.11 &1.14 
\\
\bottomrule

\end{tabular}
}
\end{table}
\setlength{\tabcolsep}{1.4pt}

\setlength{\tabcolsep}{4pt}
\begin{table}
\scriptsize
\begin{center}
\caption{Per-task fine-tuning results from~\cref{table:main_vitb} for five FGVC tasks, with a pre-trained \vit{}-B/16
}
\label{table:supp_fgvc}
\resizebox{\textwidth}{!}{
\begin{tabular}{
c l 
rrrrr r
}
\toprule
  &&\bf{\cub{}} 
  &\bf{\nabirds{}}
  &\bf{\flowers{}} &\bf{\dogs{}} &\bf{\cars{}}
  &\bf{Mean}
  \\
\midrule
\band \ttbf{(a)}&\fullft{} &87.3 &82.7 &98.8 &89.4 &84.5 &88.54
\\
\midrule
\multicolumn{4}{ l }{\emph{Head-oriented}} 
\\
\multirow{6}{*}{\ttbf{(a)}}
&\linear{} &85.3 &75.9 &97.9 &86.2 &51.3 &79.32 (0)
\\
&\partialft{}-1 &85.6 &77.8 &98.2 &85.5 &66.2 &82.63 (0)
\\
&\mlp{}-2 &85.7 &77.2 &98.2 &85.4 &54.9 &80.28 (0)
\\
&\mlp{}-3 &85.1 &77.3 &97.9 &84.9 &53.8 &79.80 (0)
\\
&\mlp{}-5 &84.2 &76.7 &97.6 &84.8 &50.2 &78.71 (0)
\\
&\mlp{}-9 &83.2 &76.0 &96.2 &83.7 &47.6 &77.31 (0)
\\
\midrule
\multicolumn{4}{ l }{\emph{Backbone-oriented}} 
\\
\multirow{6}{*}{\ttbf{(b)}}
&\sidetune{} &84.7 &75.8 &96.9 &85.8 &48.6 &78.35 (0)
\\
&\bias{} &88.4 &84.2 &98.8 &91.2 &79.4 &88.41 (3)
\\
&\adapter{}-256 &87.2 &84.3 &98.5 &89.9 &68.6 &85.70 (2)
\\
&\adapter{}-64 &87.1 &84.3 &98.5 &89.8 &68.6 &85.67 (2)
\\
&\adapter{}-8 &87.3 &84.3 &98.4 &88.8 &68.4 &85.46 (1)
\\
\midrule
\multicolumn{4}{ l }{\emph{Visual-Prompt Tuning}} 
\\
\multirow{6}{*}{\ttbf{(ours)}}&\shallowprompt{} &86.7 &78.8 &98.4 &90.7 &68.7 &84.62 (1)
 \\
 &Prompt length ($p$) &100 &50 &100 &100 &100 &90
 \\
&Tuned / Total (\%) &0.31 &0.54 &0.23 &0.20 &0.26 &0.31
\\
\cmidrule{2-8}
&\deepprompt{} &88.5 &84.2 &99.0 &90.2 &83.6 &89.11 (4)
\\
&Prompt length ($p$) &10 &50 &5 &100 &200 &73
\\
&Tuned / Total (\%) &0.29 &1.02 &0.14 &1.17 &2.27 &0.98
\\
\bottomrule
\end{tabular}
}
\end{center}
\end{table}
\setlength{\tabcolsep}{1.4pt}

\subsubsection{Numerical results of~\Cref{table:main_vitb}.}
\cref{table:supp_vtab,table:supp_fgvc} present per-task results for 24 classification tasks evaluated in~\cref{table:main_vitb}.

\subsubsection{Per-task results on training data ablations.}
\begin{figure}[t]
\centering
\includegraphics[width=0.95\textwidth]{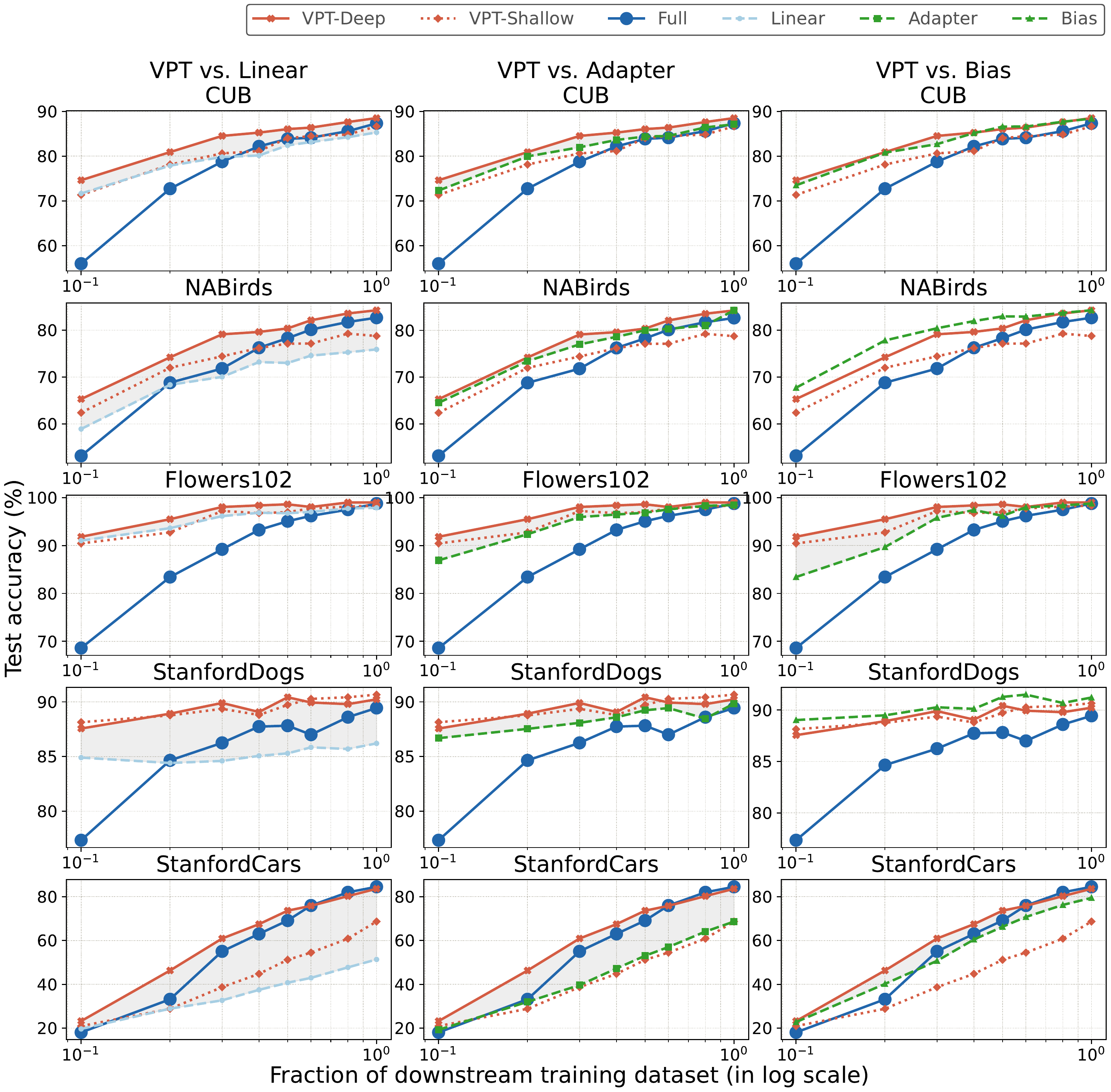}
\caption{
Effect of downstream data size, for each of FGVC tasks. The size of markers are proportional to the percentage of tunable parameters in log scale
}
\label{fig:supp_fgvcsize}
\end{figure}

\cref{fig:supp_fgvcsize} presents the per-task results for five FGVC datasets. We observe a similar trend in~\cref{fig:main_fgvcsize}: while all parameter-efficient methods outperform full fine-tuning in small-to-medium data regime, \deepprompt{} consistently surpasses \fullft{} across data scales for five FGVC tasks.

\begin{figure}
\centering
\includegraphics[width=\textwidth]{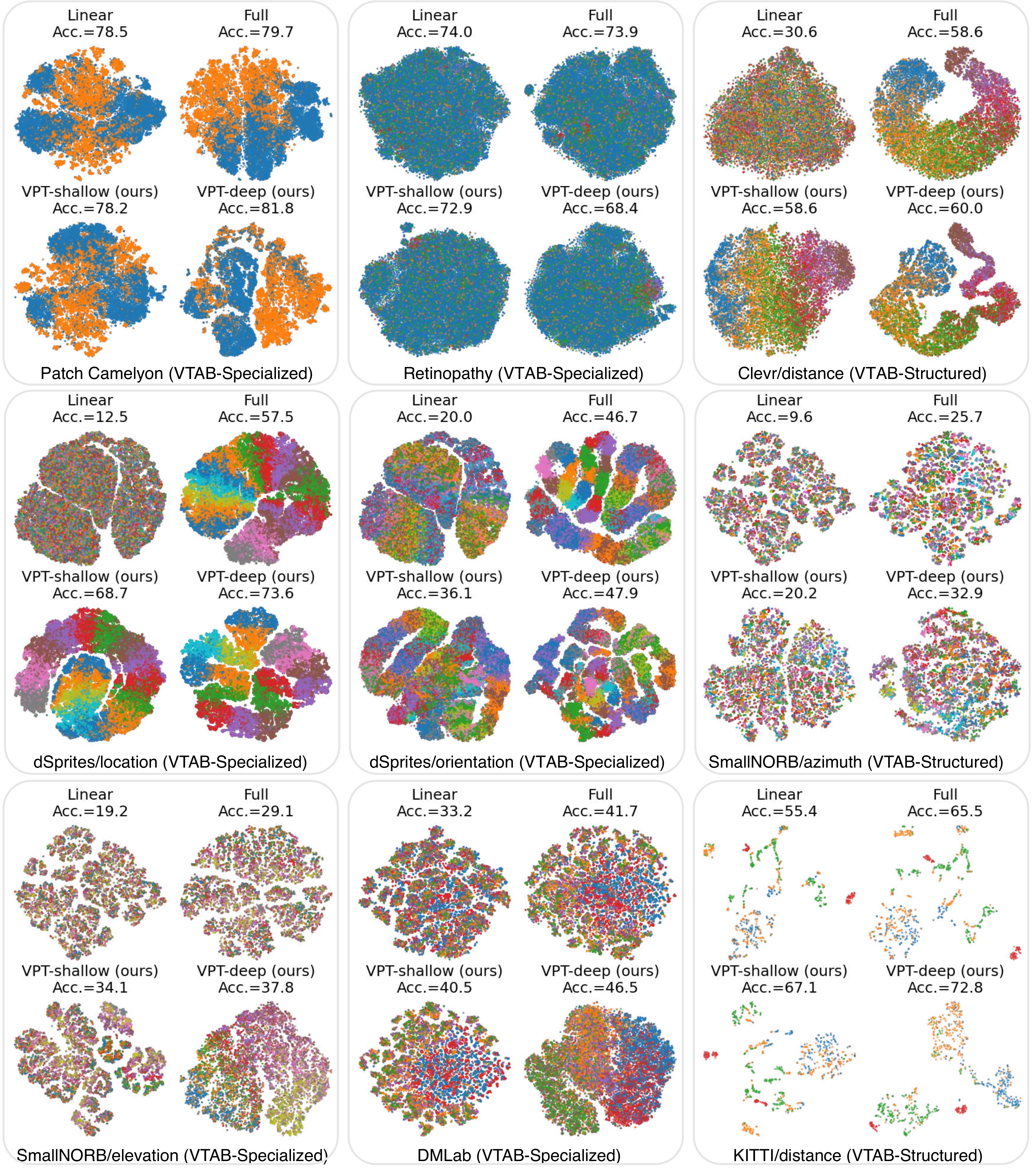}
\caption{More t-SNE visualization of the final \texttt{[CLS]} embedding $\vec{x}_N$ of more VTAB tasks. We include tasks that have less or equal to 20 target classes for visualization
}
\label{fig:more_tsne}
\end{figure}

\subsubsection{More t-SNE visualizations.}
In \cref{fig:more_tsne}, 
We presents more t-SNE visualizations, similar to~\cref{fig:tsne}, for all VTAB datasets with less than or equal to 20 target classes.

\clearpage
%
%
\bibliographystyle{splncs04}
\bibliography{egbib}

\end{document}